\documentclass{article}

\usepackage{arxiv}
\usepackage{float}
\usepackage{multirow}
\usepackage[utf8]{inputenc} 
\usepackage[T1]{fontenc}    
\usepackage{hyperref}       
\usepackage{url}            
\usepackage{booktabs}       
\usepackage{amsfonts}       
\usepackage{nicefrac}       
\usepackage{microtype}      
\usepackage{lipsum}
\usepackage{amsmath} 
\usepackage{graphicx}
\graphicspath{ {./images/} }
\usepackage{array}

\title{ConMatFormer: A Multi-attention and Transformer Integrated ConvNext based Deep Learning Model for Enhanced Diabetic Foot Ulcer Classification}

\author{
Raihan Ahamed Rifat \\
  Department of Information Technology\\
  Charles Darwin University\\
  Sydney, Australia\\
  \texttt{ahmedrifat008@gmail.com} \\
   \And
Fuyad Hasan Bhoyan \\
  Department of Computer Science and Engineering\\ University of
 Liberal Arts Bangladesh\\
  Dhaka, Bangladesh \\
  \texttt{fuyad.hasan.cse@ulab.edu.bd} \\
  \And
 Md Humaion Kabir Mehedi \\
  Department of Computer Science and Engineering\\ BRAC University\\
  Dhaka, Bangladesh  \\
  \texttt{humaion.kabir.mehedi@g.bracu.ac.bd} \\
     \And
Md Kaviul Hossain \\
  Department of Computer Science and Engineering\\ University of
 Liberal Arts Bangladesh\\
  Dhaka, Bangladesh \\
  \texttt{kaviul.hossain@ulab.edu.bd} \\
  \And
 Md. Jakir Hossen \\
  Center for Advanced Analytics (CAA),\\ COE for Artificial Intelligence,\\ Faculty of Engineering \& Technology (FET)\\ Multimedia University\\
 Melaka, Malaysia\\
  \texttt{jakir.hossen@mmu.edu.my} \\
     \And
M. F. Mridha\\
  Department of Computer Science and Engineering\\ American International University-Bangladesh\\
   Dhaka, Bangladesh \\
  \texttt{firoz.mridha@aiub.edu} \\
}

\begin{document}
\maketitle
\begin{abstract}
Diabetic foot ulcer (DFU) detection is a clinically significant yet challenging task due to the scarcity and variability of publicly available datasets. Limited annotated samples restrict the ability of conventional deep learning models to achieve robust generalization in real-world clinical scenarios. To solve these problems, we propose ConMatFormer, a new hybrid deep learning architecture that combines ConvNeXt blocks, multiple attention mechanisms convolutional block attention module (CBAM) and dual attention network (DANet), and transformer modules in a way that works together. This design facilitates the extraction of better local features and understanding of the global context, which allows us to model small skin patterns across different types of DFU very accurately. To address the class imbalance, we used data augmentation methods. A ConvNeXt block was used to obtain detailed local features in the initial stages. Subsequently, we compiled the model by adding a transformer module to enhance long-range dependency. This enabled us to pinpoint the DFU classes that were underrepresented or constituted minorities. Tests on the DS1 (DFUC2021) and DS2 (diabetic foot ulcer (DFU)) datasets showed that ConMatFormer outperformed state-of-the-art (SOTA) convolutional neural network (CNN) and Vision Transformer (ViT) models in terms of accuracy, reliability, and flexibility. The proposed method achieved an accuracy of 0.8961 and a precision of 0.9160 in a single experiment, which is a significant improvement over the current standards for classifying DFUs. In addition, by 4-fold  cross-validation, the proposed model achieved an accuracy of 0.9755 with a standard deviation of only  0.0031. We further applied explainable artificial intelligence (XAI) methods, such as Grad-CAM, Grad-CAM++, and LIME, to consistently monitor the transparency and trustworthiness of the decision-making process. These human-readable tools enhance the comprehension of the explanations and can substantially increase the practical use of our methodology. Our findings set a new benchmark for DFU classification and provide a hybrid attention transformer framework for medical image analysis.
\end{abstract}


\section{Introduction}\label{intro}
"Diabetic foot ulcers (DFUs)" are among the most severe and costly complications of diabetes mellitus, resulting in a high clinical and economic burden. This life-threatening disease represents a serious threat to public health and places a heavy financial burden on patients and the healthcare system. Compartiendo Despite the progress attained in the treatment of diabetes, early diagnosis of DFUs is still a challenge, specifically in low-resource settings. This highlights the need for rapid, accurate point-of-care tests that are available for timely intervention and treatment.
The ever-expanding population affected by diabetes has resulted in a greater worldwide burden of DFUs. "The International Diabetes Federation (IDF)" estimated the number of adults with diabetes to be 537 million in 2021, and it is expected to rise to 783 million by 2045~\cite{3}. "The World Health Organization" field office in Malaysia also documented that more than 800 million adults were diagnosed with diabetes in 2024, most of whom were from low- and middle-income countries~\cite{4}. Reuters~\cite{5} highlighted that approximately 445 million adults aged 30 years or older with diabetes remained untreated in 2022, 90\% of whom live in lower-resource settings, reflecting a stark disparity in healthcare access.

DFUs are not only prevalent but also significantly life threatening. Armstrong et al.~\cite{1} reported that 19--34\% of individuals with diabetes develop DFUs, which account for 85\% of diabetes-related lower-limb amputations. The five-year mortality rate exceeds 30\%, escalating to over 70\% in cases involving major amputations. These figures emphasize the critical importance of early detection and proactive management.

The financial aspect is equally horrifying. In the USA,  ~\cite{2} calculated the yearly cost of DFU treatment as in excess of \$9 billion to \$13 billion plus other associated diabetes costs. The health costs for diabetes soared over USD 966 billion in 2021, a 316\% increase in 15 years. For example, Medicare spending on diabetes drugs rose from \$7.7 billion in 2019 to \$35.8 billion in 2023 and is likely to be \$102 billion in 2026 (Axios, 2025~\cite{9}). Late-stage diagnosis and insufficient treatment capacity frequently exacerbate the financial burden in developing countries.

Current practice relies solely on clinical observation and manual classification for DFU identification, which is subjective and time-consuming. Breakthroughs in artificial intelligence (AI), such as deep learning and convolutional neural networks (CNNs), have demonstrated the potential to automate DFU classification, achieving high levels of accuracy and scalability. ~\cite{7} showed that deep learning models can identify subtle patterns in medical images that were previously undetected by human experts, which can lead to more accurate diagnoses.

Several studies have contributed to the advancement of this emerging field. ~\cite{2} suggested a hybrid CNN model with several separate convolutional branches. ~\cite{6} proposed XAI-FusionNet, which uses the hybrid features of DenseNet201, VGG19, and NASNetMobile, along with GradCAM, LIME, and SHAP, to increase diagnostic transparency. ~\cite{8} developed E-DFu-Net, a deep CNN architecture tailored to DFU classification, which underscores the promise of deep learning systems in clinical settings.

However, challenges remain in balancing high accuracy, architectural efficiency, and model interpretability. This motivated us to ask the following research question: Can we develop a deep learning model that achieves state-of-the-art (SOTA) accuracy and fuses convolutional and transformer paradigms for local and global feature perception? In practice, could such a model be computationally efficient and interpretable enough for clinical application?

To address these challenges, we propose a novel hybrid architecture, ConvMatformer, which integrates convolutional modules with multi-head self-attention mechanisms. The model embeds CBAM and DANet attention modules within a multistage pipeline to enhance both low- and high-level feature representations. We benchmarked ConvMatformer against six leading models, Swin-T, MaxVit-T, FastVit-MA36, ConvNextV2-T, EfficientNet-B0, and MobileNetV2, in a medical image classification task.



The major contributions of this study are outlined below.
\begin{itemize}
    \item We introduce a novel hybrid architecture, termed ConMatFormer, which ingeniously integrates ConvNext blocks with multiple attention mechanisms, including CBAM and DANet, alongside transformer modules. This architecture facilitates exceptional local feature extraction and robust global context modeling. This integration effectively captured subtle dermatological characteristics, ensuring a comprehensive representation of various types of foot ulcers.

    \item We addressed the challenge of class imbalance by employing augmentation techniques and ConvNeXt blocks in the initial stages of the model to capture detailed local features. In contrast, transformer modules in later stages enhance the comprehension of long-range contexts. This structured methodology significantly improves the diagnostic accuracy of minority and underrepresented DFU classes.
    \item We validated the robustness, accuracy, and flexibility of the proposed model by comprehensive experimental assessment using the highly recognized DS1 (DFUC2021) and DS2 (diabetic foot ulcer (DFU)) datasets. Under similar settings, our architecture outperformed the SOTA CNN and ViT models, significantly enhancing the current standards for foot ulcer categorization.
    \item We strengthen the evaluation with statistical validation (confidence intervals, cross-validation, and t-tests), ensuring the robustness and reliability of the reported results.

    \item Explainable AI methods, including Grad-CAM, Grad-CAM++, and LIME, were used to make the proposed ConMatFormer model more trustworthy and clear.
\end{itemize}

The remainder of this paper is organized as follows: Section 2 presents a detailed literature review. Section 3 presents the methodology, including the dataset description, attention modules, and proposed model. The results are presented in Section 4, and the discussion is in Section 5. In addition, the limitations and future work are presented in Section 6. Finally, we conclude the paper in Section 7. 
\section{Background Study}\label{review}

\subsection{Attention Mechanisms, Transformer-Based, and Hybrid Architectures}

Multiple convolutional neural networks (CNNs) and transformer-based models have been employed in recent research to critically evaluate deep learning (DL) algorithms for classifying diabetic foot ulcer (DFU). For instance, \cite{10} introduced a two-track scheme in which local features were obtained from the EMADN, consisting of an efficient multi-scale attention-driven network, and global features were extracted using a Swin transformer. To evaluate the proposed model, they employed the "DFUC-2021 dataset", which contains 5,955 annotated DFU images. Through the interpolation of the Swin-T and EMADN teachers, and further enhanced by Shuffle Attention, the model was able to reach best performance, having an accuracy of 78.79\% and a macro F1-score 80\%.
However, the current models have problems, including non-explainability, attention mechanism, or annotation, which motivated us to propose an interpretable remedy. \cite{11} developed Dense-ShuffleGCANet, a DL architecture for DFU classification, as well as previous work. For spatiotemporal feature extraction, they modified DenseNet-169 with a "Channel-Centric Depth-Wise Group Shuffle (CCDGS)" block along with Triplet Attention and the DFU-2021 dataset with 5955 labeled images. We see that as per the results, this method had an accuracy of 86.09\% and an F1-score of 85.77\%.
The study found that the model still failed to generalize when applied to different medical information and when utilizing the computer effectively. In another work, \cite{13} proposed DFU-SIAM, a Siamese architecture for multi-class DFU classification. The model was trained on the DFUC2021 dataset, which leveraged BEiT and EfficientNetV2S to extract both local and global features from pairs of images. To remedy the class imbalance, DFU-SIAM enriched the data in multiple ways, such as color and image transformation. 

The final predictions were made using KNN algorithm, which was modified for the macro F1-score. Although the study found that their proposed CNN outperformed traditional CNNs, it also revealed issues with real-time implementation and computational complexity. The study showed computational complexity and real-time implementation concerns even though it indicated that their suggested CNN performed better than conventional CNNs.
\cite{16} used the DFUC 2021 dataset in their investigation to look into how well CNNs and Vision Transformers performed for DFU classification. They found that while CNNs—more especially, BiT-ResNeXt50—are superior at capturing spatial characteristics, they frequently outperform Vision Transformers when data is limited. Additionally, it was noted that the Sharpness-Aware Minimization (SAM) modification improved both models. More emphasis was given on SAM’s requirement of two forward passes per update in SAM to process a signal. It increased the amount of work that must be executed by the computer system which potentially made real-time application more difficult in resource-scarce settings.
To improve the efficiency of DFU detection in medical images, a mixed DL model comprising ResNet50 and generative adversarial networks (GANs) was also suggested \cite{17}. Its ResNet50 model achieved an F1-score of 0.75 and an average accuracy and precision score of 0.76. 
When paired with GANs, the model's performance improved, yielding an F1-score, accuracy, and precision of 0.84, 0.84, and 0.85, respectively. This will affect the incorporation of GANs, which would improve DFU classification by being more advantageous for feature representation. According to the study, the blended model's added complexity would make it challenging to apply in hospital environments with little funding.
\cite{24}  created the attention-enhanced stacked parallel network (AESPNet) to improve the automatic DFU detection prediction from foot pictures. This facilitates reviewing the most current advancements. The publicly available DFUC2021 dataset was used to test the proposed model. AESPNet achieved an astounding 99\% accuracy, surpassing popular models like AlexNet, VGG16, DenseNet121, and InceptionV3. Although the attention mechanism improved the model’s performance to a great extent, our experiment demonstrated that the improvement also made it more difficult to comprehend the model’s results. New developments in hybrid AI systems have depicted a great deal of promise in solving complex diagnostic problems. For example, a thorough analysis of hybrid Machine Learning/ Deep Learning models for structural health monitoring by \cite{ref1} showed how accruing different learning paradigms might improve interpretability and resilience.
Similarly, prediction models that combine deep learning with domain-specific modelling also show improved reliability in applications similar to medical-image-based diagnostics \cite{ref2}. In order to balance accuracy and computational efficiency—two crucial factors in the building of models like ConMatFormer for DFU detection—hybrid optimization-neural architectures have been investigated concurrently \cite{ref3}. When taken as a whole, these studies highlight the growing significance of hybrid techniques, which, albeit coming from many disciplines, offer methodological support and validation for our suggested DFU classification system.

\subsection{CNN-Based, Multi-Scale, and Ensemble Learning Approaches}
\cite{12} used pre-trained CNN architectures, such as ResNet50, VGG16, VGG19 and DenseNet, to develop a transfer learning method to distinguish between diabetic foot ulcer infections and ischaemia. The “DFU2020 dataset” contains about 7000 enhanced image patches. "DFU2020 dataset" includes almost seven thousand improved image patches. With 84.76\% accuracy for infection and 99.4\% accuracy for ischemia, ResNet50 demonstrated the best performance. The study did point up certain issues, though, like the imbalanced dataset that made infection classification more difficult and the requirement for additional diverse data to enhance generalization. 
\cite{14} created DFU\_SPNet, a convolutional neural network architecture that uses stacked parallel convolutional layers with different kernel sizes to detect both local and global properties, to differentiate between DFU and normal skin. The model contained three blocks of convolutions in parallel and was trained on the DFUNet dataset using stochastic gradient descent with momentum with a learning rate of 1e-2. DFU\_SPNet outperformed other SOTA techniques with an AUC of 0.974. The study's findings suggest that parallel convolutional architectures could improve DFU classification accuracy. By creating two convolutional neural networks, DFU\_FNet and DFU\_TFNet, \cite{15} improved this field and introduced DL into the process of automatically classifying DFUs.
DFU\_FNet is a low-level method that extracts features and uses them to train the SVM and KNN classifiers. In contrast, DFU\_TFNet performs transfer learning to improve the classification. DFU\_TFNet outperformed AlexNet, GoogleNet, and VGG16 with an accuracy of 99.81\%, precision of 99.38\%, and F1-score of 99.25\%. The authors stated that although the results were positive, it could be difficult to use these models in real time on embedded systems such as FPGAs because of the limited available memory on the chip and the size of the models.
This shows that they need to be improved before they can be used in real-life clinical settings. Four hybrid CNN models were proposed in \cite{25} that used multi-branch parallel convolutional layers to distinguish between DFU and healthy skin. Each model had six blocks of parallel convolutions with different numbers of branches. The goal was to improve the feature extraction. The model with four branches performed the best, achieving an F1-score of 95.8\% on the DUF dataset. The researchers, however, stated that simply making the network deeper does not always improve performance and can even cause problems, such as gradient vanishing. Therefore, they suggested that the network design should be reconsidered. DFU-VGG was created by \cite{18} as an improved version of the VGG-19 convolutional neural network that is specifically made for DFU classification. We added batch normalization layers after each convolutional block to solve the vanishing gradient problem that arises during training. We have also changed the ReLU activation to LeakyReLU. 
The model was trained on the DFUC2021 dataset and achieved an F1-score of 86.2\%, which is superior to the original VGG19 and other SOTA CNN models. Despite the increased accuracy and generalization of the model, the study notes that the time it takes to train remains a challenge and that more work must be done before it is ready for use on a larger or real-time scale. \cite{19}, presents an ensemble approach fusing five pre-trained CNN models: VGG16, VGG19, ResNet-50, InceptionV3, and DenseNet-201. DFU were classified according to this framework. The ensemble was developed for sorting DFU and trained based on the DFUC2020 dataset. The model with a Kappa>91.85\% and an average accuracy of 95.04\% on five-fold cross-validation demonstrated good performance. The authors stated that ensemble methods can complicate a model and demand more computing power, even when they work.

\subsection{Lightweight, Thermogram-Based, and XAI-Focused Models}
\cite{20} proposed a DL method to accurately categorize DFU into four different stages. A convolutional neural network model was constructed and initially trained using 2,000 labeled images from a publicly accessible Kaggle dataset. The proposed system worked effectively for multi-class DFU stage detection, and the overall accuracy of the system was 92.6\%. However, the authors admit that the generalizability of the model to larger clinical applications could be limited by the relatively small dataset and potential class imbalance. \cite{21} presented a ML based framework which utilizes thermogram images for early detection of diabetic foot complications. Among the models tested, the conservative approach of NeuronPalm, a lightweight CNN, outperformed many conventional classifiers with an F1-score of approximately 95\%. An AdaBoost classifier was also developed with a 97\% F1-score with 10 features. The performance of the proposed framework is heavily mediated by the consistency and quality of the thermogram images and can be affected by shifts in user-dependent manipulation, device calibration, and different environmental conditions. This may introduce non-reliability in home monitoring in an uncontrolled environment. DFINET, a convolutional neural network designed to classify DFU images into infected and non-infected images, was presented by \cite{22} as a follow-up to previous work. It obtained 91.98\% accuracy and a 0.84 Matthews correlation coefficient in binary classification between ulcer and non-ulcer areas after training with the DFU image dataset, indicating high performance in distinguishing between the classes.  \cite{23} created a DL model based on the EfficientNet architecture to improve DFU detection. A dataset of 844 foot photos, comprising both healthy and ulcerated cases, was used to train the framework. The dataset for training the approach included 844 photos of the foot, either healthy or with ulceration. EfficientNet achieved 98.97\% accuracy, which is higher than fashion related benchmarks -AlexNet, GoogleNet, VGG16, and VGG19. It also had an F1-score of 98\%, recall of 98\%, precision of 99\%. The model’s performance will be a function of the quality and diversity of the training data, and to show its efficacy in the future, it will need to be validated on larger and more diverse datasets. \cite{6} proposed an XAI-based scheme for detecting DFUs, called XAI-FusionNet, which integrates XAI methodologies with multi-modal feature fusion. This framework uses higher frequency features in the input images using pre-trained CNNs, such as NASNetMobile, VGG19, and DensNet201. To further improve the classification accuracy of these features, we applied a meta-tuner module. For instance, after training with 6963 skin images, XAI-FusionNet achieved an accuracy of 99.05\%, recall of 98.18\%, precision of 100\%, AUC of 99.09\%, and F1-score of 99.08\%. The model uses XAI techniques, such as SHAP, LIME, and GradCAM, to make interpretable predictions. To classify DFUs into infection and ischaemia, \cite{7} also developed a DL model based on the EfficientNet. This model, trained on an augmented dataset of DFU images, achieved 99\% accuracy for ischaemia classification and 98\% accuracy for infection classification. They had better results than prior models such as ResNet and Inception, with a top 1 accuracy of 87\%. Furthermore an ensemble CNN reached 90\% for ischaemia classification and 73\% for infection classification. The study was hampered by a small and limited dataset, the use of expert-labeled images, and inadequate testing in real clinical circumstances.

\section{Methodology}\label{method}
\begin{figure}[htbp]
    \centering
    \includegraphics[scale=.45]{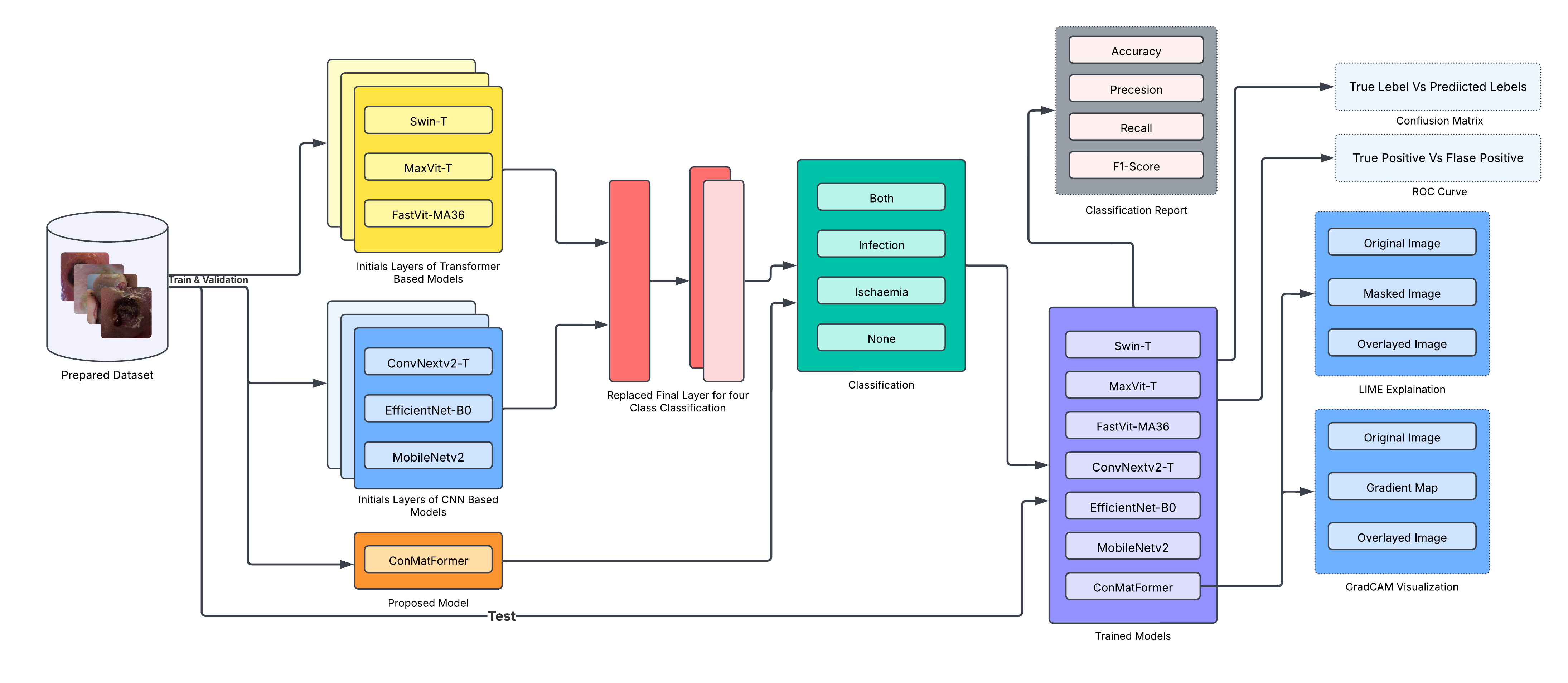}
    \caption{Workflow Diagram of the Overall Experiment of DFU Classification.}
    \label{fig:dfu_overall}
\end{figure}

The experimental workflow for diabetic foot ulcer classification involved training and testing multiple baseline models, including transformer- and CNN-based models, as illustrated in Figure~\ref{fig:dfu_overall}. In addition, the hybrid model, ConMatFormer, was evaluated in the same settings and environment. After training, the models were assessed using evaluation metrics such as accuracy, precision, recall, F1-score, confusion matrices, and ROC curves. Moreover, explainability was addressed using the LIME and Grad-CAM techniques.

\subsection{Dataset Description}
\begin{figure}[htbp]
    \centering
    \includegraphics[width=0.75\textwidth]{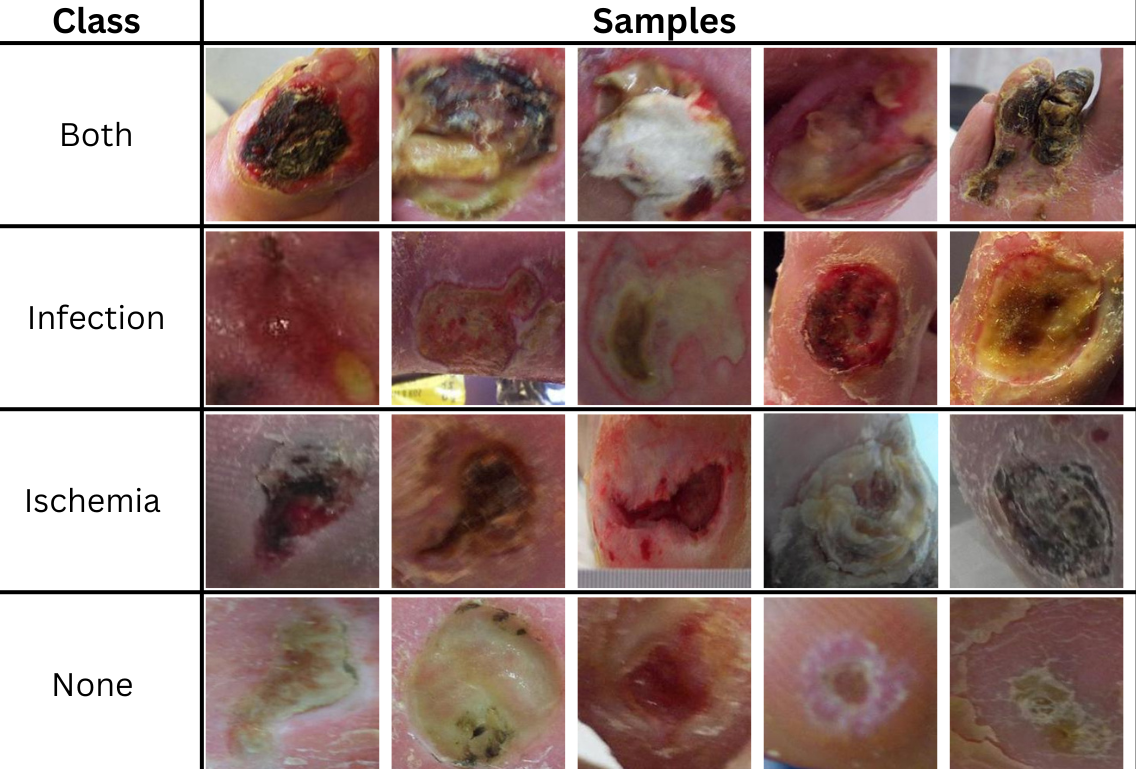}
    \caption{Sample Images from DS1 (DFUC2021) Datasets.}
    \label{fig:dfu1}
\end{figure}

The "DFUC2021" \cite{dfu2021} utilized in this study, as illustrated in Figure~\ref{fig:dfu1}. The DFU-2021 Grand Challenge Dataset consists of a substantial collection of 15,683 patch images, with 5,955 allocated for training, 5,734 for testing, and 3,994 remaining unlabeled. Each patch was classified into four categories: none, infection, ischemia, and both (infection and ischemia). Images were collected from the Lancashire Teaching Hospital over the past few years. The dataset was divided into training, validation, and test subsets and was extensively employed for multi-class classification and supervised learning tasks. The DS1 data patches were derived from high-resolution clinical foot images and meticulously curated to facilitate robust model training for practical diagnosis. 

\begin{figure}[htbp]
    \centering
    \includegraphics[width=0.75\textwidth]{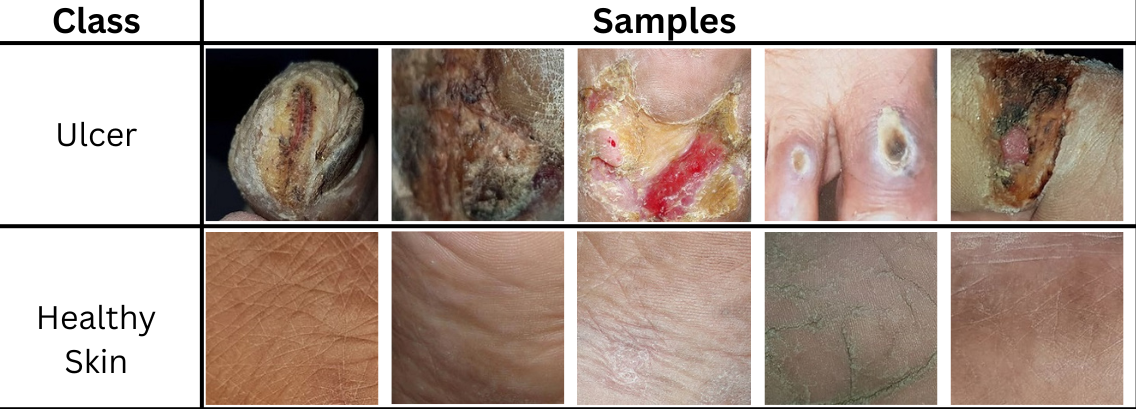}
    \caption{Sample Images from Kaggle DS2 (diabetic foot ulcer (DFU)) Dataset by Laithi.}
    \label{fig:dfu2}
\end{figure}

In contrast, the second dataset we used, "diabetic foot ulcer (DFU)" or "DS2" taken from Kaggle \cite{dfu}, shown in Figure~\ref{fig:dfu2} offers a smaller and more refined binary classification set. DS2 originates from the Nasiriyah Hospital’s diabetic center (southern Iraq). Out of the total 1055 marked patches that make up the collection, 543 are deemed healthy, while 512 are noted to be unhealthy. Each patch in this dataset was made from 754 full foot, (224 x 224) pixel images that were captured with electronic devices like tablets and cellphones and categorized as either ulcerated (DFU) or normal skin. \\

Collectively, these two datasets provide a comprehensive approach to DFU classification, employing multiple machine learning (ML) and deep learning (DL) models. From basic to fine-grained pathology analyses, these datasets have the potential to make significant contributions. Although the datasets complement each other by addressing different complexities, utilizing both datasets enables the models to learn from high-quality labels.

\subsection{Data Processing}
DS1 and DS2, the datasets used in this investigation, are made up of labelled data with several classes. Both datasets initially showed a class imbalance, with 227 samples in DS1 being categorized as "ischemia" and 2,652 samples in the "none" (no pathology) group. Both datasets were enhanced in order to rectify this imbalance and improve the model's generalizability; this led to a balanced post-augmentation distribution of samples per class in the binary classification subset and multiclass dataset. Prior to augmentation, a stratified approach was used to separate the datasets into training, validation, and test sets. 60\% of the samples in the multi-class dataset were used for training, 20\% for validation, and 20\% for testing. The test samples were left unaltered while the training and validation samples underwent augmentation. The binary classification subset was treated in the same way, yielding 700 samples for both classes in the training subset and 200 samples for both classes in the validation subset following augmentation. By employing modifications like random image flipping, both horizontally and vertically, data augmentation improved the datasets’ diversity. Moreover, moving the images at different angles enhanced the models’ ability to recognize images from various perspectives.  These augmentations significantly mitigated overfitting and improved the generalization capabilities of the models, as shown in Table~\ref{tab:data-distribution}.

\begin{table}[ht]
\centering
\caption{Dataset details, sample balancing, and data split.}
\label{tab:data-distribution}
\begin{tabular}{|c|c|ccc|ccc|}
\hline
\textbf{Dataset} & \textbf{Class} &
\multicolumn{3}{c|}{\textbf{Before Augmentation}} &
\multicolumn{3}{c|}{\textbf{After Augmentation}} \\
\cline{3-8}
 & & \textbf{Train} & \textbf{Validation} & \textbf{Test} &
     \textbf{Train} & \textbf{Validation} & \textbf{Test} \\
\hline
\multirow{4}{*}{DFU 2021} 
 & None (no pathology) & 1630 & 511 & 511 & 3036 & 1012 & 511 \\
 & Infection           & 1533 & 511 & 511 & 3035 & 1011 & 511 \\
 & Ischaemia           & 135  & 46 & 46 & 2777 & 925  & 46 \\
 & Both                & 370  & 125 & 125 & 3034  & 1011 & 125 \\
\hline
\multirow{2}{*}{DFU} 
 & Ulcer (Abnormal)    & 312  & 100 & 100 & 700 & 200 & 100 \\
 & Healthy Skin (Normal) & 343 & 100 & 100 & 700 & 200 & 100 \\
\hline
\end{tabular}
\end{table}

\begin{table}[h]
\centering
\caption{Image augmentation strategies used in this study (input size: 224$\times$224)}
\label{tab:augmentation}
\begin{tabular}{|l|l|}
\hline
\textbf{Transformation} & \textbf{Description / Parameters} \\
\hline
Resize & Resize images to $224 \times 224$ \\
\hline
Random Horizontal Flip & Flip horizontally with probability $p \in \{0.2, 0.5\}$ \\
\hline
Random Vertical Flip & Flip vertically with probability $p \in \{0.2, 0.5\}$ \\
\hline
Random Rotation & Rotate by a random degree from $\{15^\circ, 30^\circ, 45^\circ, 60^\circ\}$ \\
\hline
Random Affine & Apply affine transform with parameters: \\
              & \quad Rotation $= 10^\circ$ \\
              & \quad Translation $= (0.1, 0.1)$ \\
              & \quad Scale range $= (0.9, 1.1)$ \\
\hline
\end{tabular}
\end{table}

To apply augmentation, each image was first resized to 224 x 224 pixels, so that all inputs shared a consistent resolution, as shown in Table \ref{tab:augmentation}. After resizing, the images were randomly flipped both horizontally and vertically with probabilities randomly chosen between 0.2 and 0.5, which assisted the neural network in learning the orientation-invariant of the images. To enable the models to cope with various image capture angles, random rotations with rotation angles of 15, 30, 45, and 60 degrees were performed. Finally, a random affine transformation was applied, which enabled up to $\pm 10$ degree of additional rotation, up to 10\% translation in both the x and y directions, and a scale change between 0.9 and 1.1. These augmentations produced diverse training samples for our experiments which helped in reducing overfitting and aided the model generalize better to unseen DFU images.

\subsection{Environmental Setup and Model Selection}

\begin{table}[h]
\centering
\caption{Training Configuration and Hardware Specifications.}
\label{tab:training-config}
\begin{tabular}{|l|l|}
\hline
\textbf{Parameter} & \textbf{Value} \\
\hline
Epochs & 50 \\
Batch Size & 64 \\
Optimizer & Adam \\
Learning Rate & 0.00001 \\
Weight Decay & 0.00003 \\
\hline
\textbf{Hardware} & \textbf{Specification} \\
\hline
Processor & 12-core, 24-thread CPU \\
GPU & NVIDIA RTX 4070 Ti Super (16GB) \\
RAM & 64 GB \\
\hline
\end{tabular}
\end{table}

We executed all our programs in the local environment. Table~\ref{tab:training-config} shows the hardware and training settings that were used for the experiments. Using the Adam optimizer, the training was performed over 50 epochs with a batch size of 64. We chose a learning rate of 0.00001 and a weight decay of 0.00003 to ensure that convergence was stable and that we did not overfit.

We chose a variety of models for comparison to determine the performance of different deep learning models. Some of these include transformer-based architectures such as Swin-T, MaxViT-T, and FastViT-MA36, and convolutional backbones such as ConvNeXtV2-T, EfficientNet-B0, and MobileNetV2. In addition, the proposed hybrid model, ConMatFormer, combines the best features of CNNs and transformers. All models were trained and tested on a high-performance workstation with a 12-core, 24-thread processor, 64 GB of RAM, and an NVIDIA RTX 4070 Ti Super GPU with 16 GB of VRAM.

\subsection{Attention mechanism}
The attention mechanism allows a computer to focus on specific parts of an input in a manner similar to human visual focus. Attention mechanisms aim to enhance the understanding of features by CNNs via adaptive weights in terms of various regions or channels. This mechanism enables the network to focus on significant region(s), such as lesions in medical images or objects involved in scene recognition. To determine what and where to focus on in feature maps, deep CNNs use popular attention techniques like squeeze-and-excitation (SE) and the convolutional block attention module (CBAM). Compared to SOTA approaches, this approach produces a better model for intricate visual tasks.
\subsubsection{Convolutional Block Attention Module (CBAM)}

The convolutional block attention module, introduced by \cite{26}, is an effective attention mechanism designed to enhance the representational power of intermediate feature maps \( F \in \mathbb{R}^{C \times H \times W} \). Each CBAM block consists of two sequential attention sub-modules: \textit{channel attention} and \textit{spatial attention}, which generate a 1D channel attention map \( M_c \in \mathbb{R}^{C \times 1 \times 1} \) and a 2D spatial attention map \( M_s \in \mathbb{R}^{1 \times H \times W} \), respectively, as follows: These attention maps are applied to the input features via element-wise multiplication to produce a refined output \( F'' \).

In the \textbf{channel attention} module, feature descriptors are obtained using both average and max pooling operations along the spatial dimensions. These descriptors are then passed through shared multi-layer perceptrons (MLPs), denoted as \( W_0 \) and \( W_1 \), followed by a sigmoid activation function to compute channel-wise attention.

The \textbf{spatial attention} module further refines the feature representation by applying average and max pooling along the channel axis, generating intermediate maps \( F_{s\_avg} \) and \( F_{s\_max} \). These are concatenated and processed through a convolutional layer with a large kernel size (typically \( 7 \times 7 \)) to capture the contextual spatial dependencies, thereby producing a spatial attention map.

The CBAM enhances the network's focus on salient features, improving both the discriminative power and generalization capability.

\subsubsection{Dual Attention Network (DANet)}

The dual attention network is a DL architecture developed to enhance feature representation \cite{danet}, particularly in semantic segmentation, but it is also applicable to classification tasks. DANet integrates two complementary attention mechanisms—\textit{position attention} and \textit{channel attention}—to jointly model spatial and channel-wise dependencies.

\textbf{Position Attention Module (PAM)}:  
The Position Attention Module captures long-range contextual information by computing attention across spatial positions. Given a local feature map \( A \in \mathbb{R}^{C \times H \times W} \), it is passed through two parallel convolutional layers to produce two feature maps \( B \) and \( C \), each with dimensions \( C \times H \times W \). These are reshaped into matrices of size \( C \times N \), where \( N = H \times W \) is the number of spatial positions. A spatial attention map \( S \in \mathbb{R}^{N \times N} \) is then computed using dot product similarity, followed by a softmax operation, as shown in Equation~\ref{eq:pam_softmax}.

\begin{equation}
s_{ji} = \frac{\exp(B_i \cdot C_j)}{\sum_{i=1}^{N} \exp(B_i \cdot C_j)}
\label{eq:pam_softmax}
\end{equation}

Meanwhile, the original input \( A \) is processed through a third convolutional layer to obtain \( D \in \mathbb{R}^{C \times H \times W} \), which is reshaped to \( C \times N \). This is multiplied by the transpose of \( S \), and the result is reshaped back to \( C \times H \times W \). The final output is computed as follows Equation~\ref{eq:pam_output}:

\begin{equation}
E_j = \alpha \sum_{i=1}^{N} (s_{ji} D_i) + A_j
\label{eq:pam_output}
\end{equation}

where \( \alpha \) is a learnable scaling parameter. The PAM enables spatially distant yet semantically similar features to interact, enhancing the model’s ability to capture the global context.

\textbf{Channel Attention Module (CAM)}

The Channel Attention Module (CAM) focuses on inter-channel relationships. Starting with the input feature map \( A \), it is reshaped to \( C \times N \) and multiplied by its transpose to form a similarity matrix across channels. A softmax operation is applied to generate the channel attention map \( X \in \mathbb{R}^{C \times C} \) as shown in Equation~\ref{eq:cam_softmax}.

\begin{equation}
x_{ji} = \frac{\exp(A_i \cdot A_j)}{\sum_{i=1}^{C} \exp(A_i \cdot A_j)}
\label{eq:cam_softmax}
\end{equation}

The attention-refined feature map is then obtained as follows Equation~\ref{eq:cam_output}:

\begin{equation}
E_j = \beta \sum_{i=1}^{C} (x_{ji} A_i) + A_j
\label{eq:cam_output}
\end{equation}

where \( \beta \) is a learnable scaling parameter. The CAM enhances feature representation by emphasizing important channels and suppressing less-relevant channels.

\subsubsection{ConvNextV2 block}

By combining advancements from contemporary vision transformers and an effective CNN design, the ConvNeXtV2 block is an improved convolutional architecture that expands upon \cite{onvNextV2}. It introduces advances including depth wise convolutions, layer normalization, activation scaling, and stability-enhancing approaches while maintaining the design philosophy of deep convolutional networks.
.

\textbf{Depthwise Convolution:}  
Given an input tensor \( X \in \mathbb{R}^{C \times H \times W} \), a depthwise convolution is first applied to each channel independently, as shown in Equation~\ref{eq:dwconv}.
\begin{equation}
X' = \text{DWConv}(X)
\label{eq:dwconv}
\end{equation}

\textbf{Layer Normalization (Channel-last):}  
After depthwise convolution, Layer Normalization is applied to stabilize training Equation~\ref{eq:layernorm}:
\begin{equation}
\hat{X}_{i,j,k} = \frac{X_{i,j,k} - \mu_i}{\sqrt{\sigma_i^2 + \epsilon}} \cdot \gamma + \beta
\label{eq:layernorm}
\end{equation}
where \( \mu_i \) and \( \sigma_i^2 \) are the mean and variance computed over the spatial dimensions, and \( \gamma \) and \( \beta \) are learnable parameters.

\textbf{Pointwise Convolutions with GELU Activation:}  
Two pointwise (1x1) convolutions are applied with a GELU nonlinearity in between Equation~\ref{eq:pwconv}:
\begin{equation}
X'' = W_2 \cdot \text{GELU}(W_1 \cdot \hat{X})
\label{eq:pwconv}
\end{equation}
where \( W_1 \in \mathbb{R}^{C \times 4C} \) and \( W_2 \in \mathbb{R}^{4C \times C} \) are the learnable projection matrices.

\textbf{Residual Connection with Scaling:}  
A residual connection is added with a learnable scaling factor \( \alpha \) (e.g., initialized as \(1e{-6}\)) Equation~\ref{eq:residual}:
\begin{equation}
Y = X + \alpha \cdot X''
\label{eq:residual}
\end{equation}

The complete ConvNeXtV2 block can be expressed as Equation~\ref{eq:convnextv2}:
\begin{equation}
Y = X + \alpha \cdot W_2 \cdot \text{GELU}(W_1 \cdot \text{LayerNorm}(\text{DWConv}(X)))
\label{eq:convnextv2}
\end{equation}

This architecture efficiently combines the strengths of convolutional operations and transformer-inspired components, resulting in a powerful and scalable building block for vision-related tasks.

\subsection{Proposed ConvMatformer}
\begin{figure}[htbp]
    \centering
    \includegraphics[width=\textwidth]{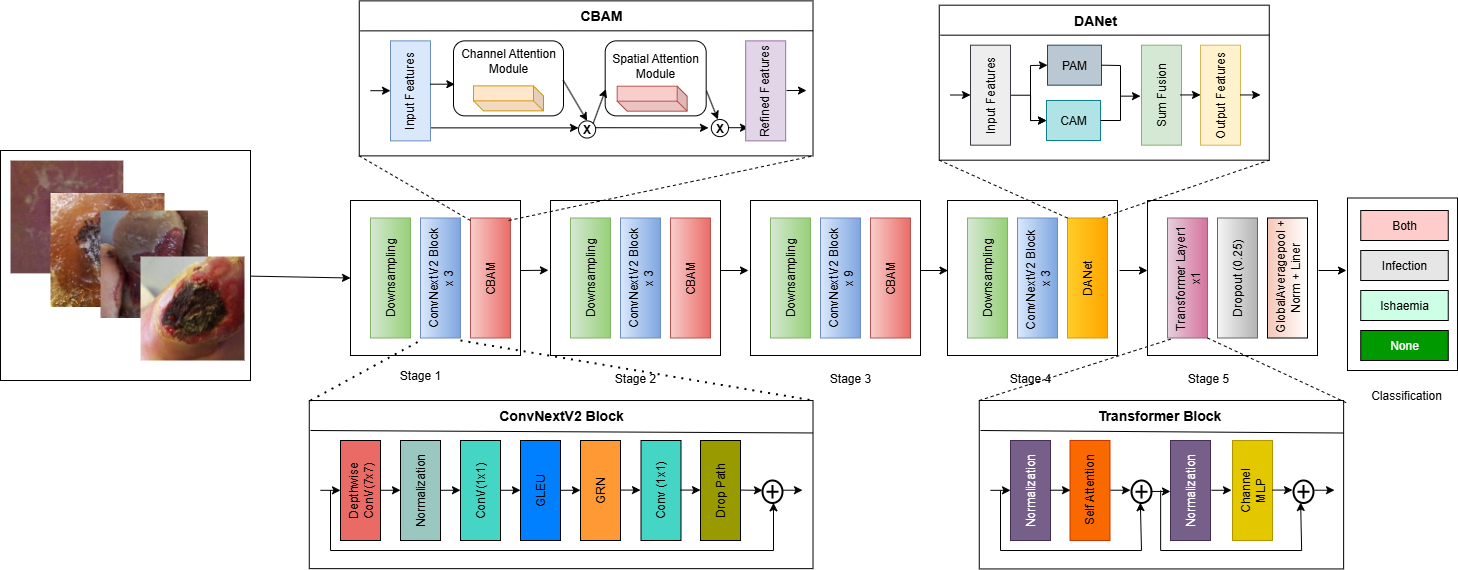}
    \caption{Architecture of the proposed ConMatFormer integrating ConvNeXtV2, CBAM, DANet, and Transformer for DFU classification.}
    \label{fig:ConMatFormer_architecture}
\end{figure}

The ConvMatformer architecture shown in Figure~\ref{fig:ConMatFormer_architecture} introduces an innovative DL framework that integrates convolutional and transformer-based components for the effective classification of medical-image data. 
This model uses a number of hierarchical steps to gradually extract characteristics from input photos that are 224 x 224 in size. ConvNeXtV2 blocks were used for effective localized feature extraction after the CBAM was used to improve spatial and channel attention. As the network develops, it integrates transformer blocks and DANet modules to capture contextual semantics and global dependencies. Each of the five main steps of the design includes down sampling and sophisticated convolutional operation. In the last stage, global average pooling and a fully connected multi-layer perceptron (MLP) head for classification purposes are the results of attention-based fusion and transformer layers. Four categories were covered by the predictions the model produced: both, infection, ischemia, and none. The ConvMatformer exhibits remarkable ability in tackling the difficulties of medical picture interpretation by utilizing both local and global feature representations.

\begin{table}[h]
\centering
\caption{Stage-wise architecture of the proposed ConMatFormer for DFU classification.}
\label{tab:conmat_architecture}
\scriptsize
\renewcommand{\arraystretch}{1.3}
\setlength{\tabcolsep}{5pt}

\begin{tabular}{|c|c|c|c|c|}
\hline
\textbf{Stage} & \textbf{Block / Module} & \textbf{Input Shape} & \textbf{Output Shape} & \textbf{Details / Params} \\ \hline

Stem & Conv (7$\times$7), LayerNorm & [3, 224, 224] & [96, 56, 56] & 4,704 + 192 \\ \hline

Stage 1 & ConvNeXtV2 Block $\times$3 + CBAM & [96, 56, 56] & [96, 56, 56] & 239,904 + 2,412 \\ \hline

Stage 2 & ConvNeXtV2 Block $\times$3 + CBAM & [96, 56, 56] & [192, 28, 28] & 922,176 + 9,432 \\ \hline

Stage 3 & ConvNeXtV2 Block $\times$2 + CBAM & [192, 28, 28] & [384, 14, 14] & 10,841,472 + 37,296 \\ \hline

Stage 4 & ConvNeXtV2 Block + DANet & [384, 14, 14] & [768, 7, 7] & 14,305,536 + 1,248,032 \\ \hline

Stage 5 & Transformer Block + LN + Dropout & [768, 7, 7] & [768, 7, 7] & 2,362,368 + 4,722,432 \\ \hline

Pooling & Global AvgPool + LayerNorm & [768, 7, 7] & [768] & 1,536 \\ \hline

Classifier & Linear Head & [768] & [4] & 3,076 \\ \hline

\textbf{Total} & -- & -- & -- & \textbf{36.33M params, 391.95M MACs} \\ \hline
\end{tabular}

\end{table}

The suggested ConMatFormer design is thoroughly broken out step-by-step in Table \ref{tab:conmat_architecture}, guaranteeing total reproducibility. The network starts with a stem layer that uses LayerNorm and a 7 x 7 convolution to compress the input images to [96, 56, 56]. Using stacked ConvNeXtV2 blocks, stages 1 through 3 gradually improved the features. CBAM modules were added at the end of each stage to improve channel-wise and spatial attention. A ConvNeXtV2 block and a DANet module that adds position and channel attention for contextual representation are used in Stage 4 to further enhance and expand the feature maps. Global dependency modelling is made possible by Stage 5's inclusion of a local transformer block with layer normalisation and dropout.

\subsection{Interpretation with XAI}

Explainable artificial intelligence refers to a set of tools and techniques that make the decision-making processes of AI models transparent, interpretable, and understandable to humans. In complex models, especially deep neural networks, XAI helps users understand why a particular prediction was made, which features influenced the outcome, and how reliable the model’s decision is.

In medical imaging, AI models are increasingly used to diagnose diseases using radiographs, MRIs, CT scans, histopathological images, and skin lesion photographs. However, medical applications demand high accountability, trust, and interpretability because AI in clinical decisions significantly impacts human lives, and doctors must validate these against established patterns. XAI aids in trust building, debunking biases, validating predictions against clinical knowledge, and promoting adoption by nontechnical stakeholders.

LIME explains the predictions of any classifier by learning an interpretable model locally around a prediction \cite{lime}. Given an image \( x \), LIME perturbs the input by masking segments (superpixels), obtains predictions for these perturbed samples, and fits a simple interpretable model (e.g., linear regression) to approximate the behavior of the complex model \( f \) locally.

\paragraph{Key Steps:}
\begin{enumerate}
    \item Segment image \( x \) into superpixels.
    \item Generate perturbed samples by randomly turning superpixels on/off.
    \item Get predictions \( f(x') \) from the black-box model.
    \item Fit a weighted interpretable model \( g(x') \) by minimizing Equation~\ref{eq:lime}.
\end{enumerate}

\begin{equation}
    \underset{g}{\arg\min} \; L(f, g, \pi_x) + \Omega(g)
    \label{eq:lime}
\end{equation}

Where:
\begin{itemize}
    \item \( L \): Local loss between \( f \) and \( g \)
    \item \( \pi_x \): Locality measure (e.g., distance between \( x \) and \( x' \))
    \item \( \Omega(g) \): Complexity of the interpretable model
\end{itemize}

LIME highlights the regions of the input that have the highest impact on the model’s decision.\\

Grad-CAM is a class-discriminative localization technique that uses the gradients of a target class flowing into the last convolutional layer to generate a heatmap of important regions \cite{gradcam}.

Given an image \( I \) and a target class score \( y^c \), Grad-CAM computes the gradient of \( y^c \) with respect to the feature maps \( A^k \) of the convolutional layer as follows Equation~\ref{eq:gradcam-weights}:

\begin{equation}
    \alpha_k^c = \frac{1}{Z} \sum_i \sum_j \frac{\partial y^c}{\partial A_{ij}^k}
    \label{eq:gradcam-weights}
\end{equation}

Where:
\begin{itemize}
    \item \( \alpha_k^c \): Importance weight for feature map \( k \)
    \item \( Z \): Total number of pixels in \( A^k \)
\end{itemize}

Subsequently, the Grad-CAM heatmap is computed using Equation~\ref{eq:gradcam-map}:

\begin{equation}
    L_{\text{Grad-CAM}}^c = \text{ReLU} \left( \sum_k \alpha_k^c A^k \right)
    \label{eq:gradcam-map}
\end{equation}

ReLU is used to focus only on the features that positively influence the class score. The heatmap was then upsampled and overlaid on the original image for visual interpretation.

\section{Result and Analysis}
To guarantee thorough validation, all models were trained and evaluated on two benchmark datasets. To give both quantitative and qualitative insights, we used a variety of performance criteria for evaluation, including accuracy, precision, recall, F1-score, ROC curves, and confusion matrices. Two different experimental settings were used for the evaluation: a more demanding 4-fold cross validation that verified the models' stability and durability across several data divisions, and a rational data split (standard train test division). Additionally, explainable AI (XAI) tools including Grad-CAM, Grad-CAM++, and LIME were used to visualize the decision-making process in order to improve transparency and interpretability. This allowed for human-readable explanations of the learned features and model predictions. This combination of multi-metric evaluation, diverse validation strategies, and explainability analysis provides a reliable and trustworthy assessment of the proposed framework compared to state-of-the-art baselines.
\subsection{Evaluation with Rational Split}

As indicated in Table ~\ref{tab:model_performance}, the suggested ConMatFormer model fared better than every other architecture across all evaluation metrics. With the maximum accuracy of 0.8961, it demonstrated excellent predictive performance all around. It also had the best precision of 0.9160, which demonstrated how well it could reduce false positives. Its recall/sensitivity and F1-score were higher than those of the comparison models employed in this investigation, coming in at 0.8866 and 0.9004, respectively. Competitive performance was also demonstrated by models like Swin-T and FastVit-MA36, which had respective F1-scores of 0.8878 and 0.8804. The accuracy ratings of lightweight models like MobileNetV2 and EfficientNet-B0 were 0.0839 and 0.0713 lower, respectively, than those of the suggested ConMatFormer. This could be because of their restricted ability to extract complicated features.
\begin{table}[h]
\centering
\caption{Comparison of model performance on the test set on DS1 dataset.}
\label{tab:model_performance}
\begin{tabular}{|l|c|c|c|c|}
\hline
\textbf{Model} & \textbf{Test Accuracy} & \textbf{Precision} & \textbf{Recall} & \textbf{F1-Score} \\
\hline
Swin-T                 & 0.8860 & 0.8841 & 0.8848 & 0.8840 \\
MaxVit-T              & 0.8734 & 0.8895 & 0.8601 & 0.8733 \\
FastVit-MA36     & 0.8826 & 0.9109 & 0.8700 & 0.8878 \\
ConvNeXtV2-T     & 0.8860 & 0.8841 & 0.8848 & 0.8840 \\
EfficientNet-B0     & 0.8248 & 0.8240 & 0.8390 & 0.8306 \\
MobileNetV2         & 0.8122 & 0.8229 & 0.8050 & 0.8129 \\
\textbf{ConMatFormer (Proposed)} & \textbf{0.8961} & \textbf{0.9160} & \textbf{0.8866} & \textbf{0.9004} \\
\hline
\end{tabular}
\end{table}
\begin{figure}[htbp]
    \centering
    \includegraphics[width=0.9\textwidth]{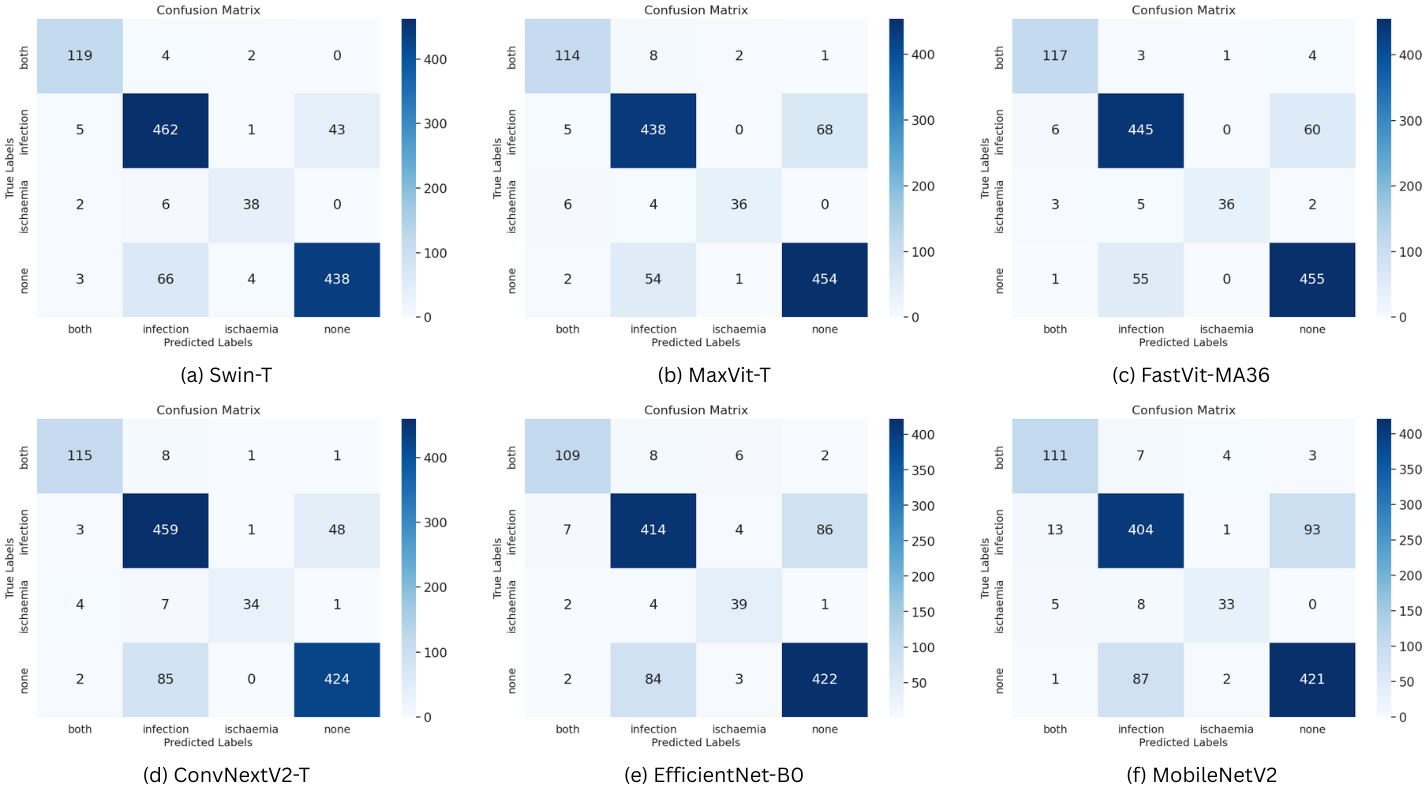}
    \caption{Confusion Matrix of Experimental Models used in this Study on DS1.}
    \label{fig:cm_dfu}
\end{figure}

To comprehensively evaluate model performance, we analyzed the confusion matrices of six state-of-the-art architectures on the DFU classification task, as illustrated in Figure~\ref{fig:cm_dfu}. Swin-T correctly predicted 119 both, 462 infection, 38 ischaemia, and 438 none cases. MaxVit-T classified 114 both, 438 infection, 36 ischaemia, and 454 none cases correctly. FastVit-MA36 demonstrated consistently solid performance by obtaining 445 infections and 455 non-cases right, along with decent results in both ischaemia classes. ConvNextV2-T matched Swin-T infection prediction but lagged slightly behind in all cases. EfficientNet-B0’s performance decreased across all categories, correctly identifying 414 infections and 422 non-cases, while struggling more with ischaemia. MobileNetV2 showed the weakest results overall, with lower counts across the board (404 for infection and 421 for none), indicating difficulty in capturing the more nuanced features of the data. Among the models, FastVit-MA36 was notable for its stability and minimal confusion between classes.\\

\begin{figure}[htbp]
    \centering
    \includegraphics[width=0.9\textwidth]{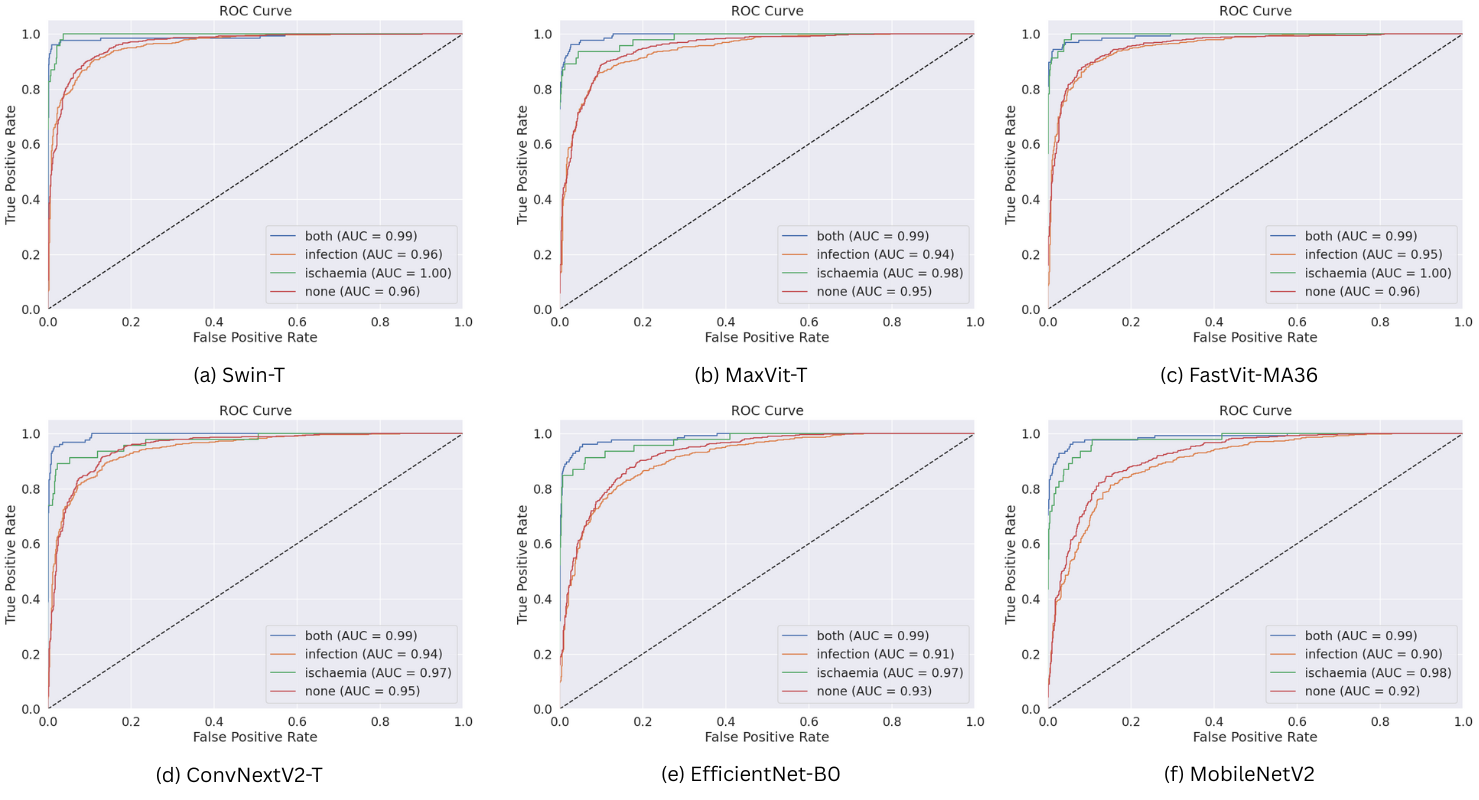}
    \caption{ROC Curve of Experimental Models used in this Study on DS1.}
    \label{fig:roc_dfu}
\end{figure}

While the confusion matrices highlight detailed class-level prediction strengths and weaknesses, Figure~\ref{fig:roc_dfu} further illustrates each model’s ability to discriminate between classes through ROC curves and AUC scores. All models achieved excellent AUC scores for the ``both” class, consistently at 0.99. Swin-T and FastVit-MA36 had the highest AUCs overall, both achieving 1.00 for ischaemia and 0.96 for non-ischaemia. MaxVit-T and ConvNextV2-T also performed well, with ischaemia AUCs of 0.98 and 0.97, respectively. EfficientNet-B0 and MobileNetV2 showed noticeably lower AUCs across all classes, especially in the infection and pneumonia categories. However, MobileNetV2 dropped to 0.90 and 0.92. These curves suggest that while most models can effectively separate classes, Swin-T and FastVit-MA36 maintain stronger and more consistent discrimination across all categories.\\

\begin{figure}[htbp]
    \centering
    \includegraphics[width=0.9\textwidth]{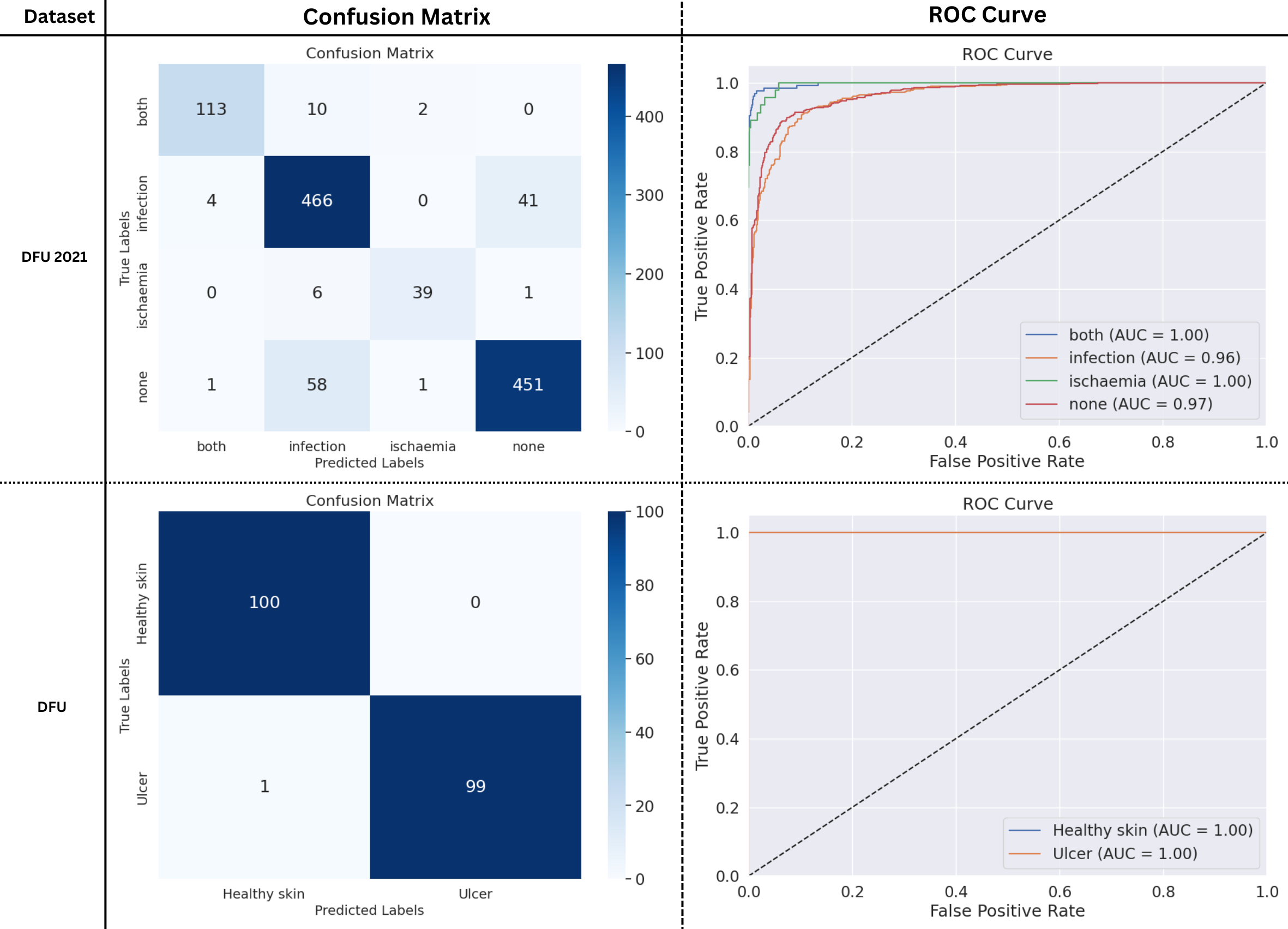}
    \caption{Confusion Matrix and ROC curve of Experimental Models used in this Study on DS1 and DS2.}
    \label{fig:cm_dfu_both}
\end{figure}

Figure~\ref{fig:cm_dfu_both} demonstrates the performance of the proposed ConMatFormer Model across the two datasets, DS1 and DS2. On the DFU2021 dataset, the model correctly classified 113 both, 466 infection, 39 ischaemia, and 451 no cases. A few misclassifications occurred, notably 58 non-image predictions as infections. The ROC curves indicated a high classification capability, with AUC scores reaching 1.00 for both and ischaemia. The infection class had an AUC of 0.96, while the non-infection class reached 0.97. In the simpler DS2 dataset, the model showed near-perfect results by classifying all 100 healthy skin samples correctly and misclassifying only one ulcer image out of 100 images. The ROC curves for this binary task reached 1.00 for both classes, reflecting the model’s ability to distinguish between healthy skin and ulceration.

\begin{figure}[h]
    \centering
    \includegraphics[width=0.9\linewidth]{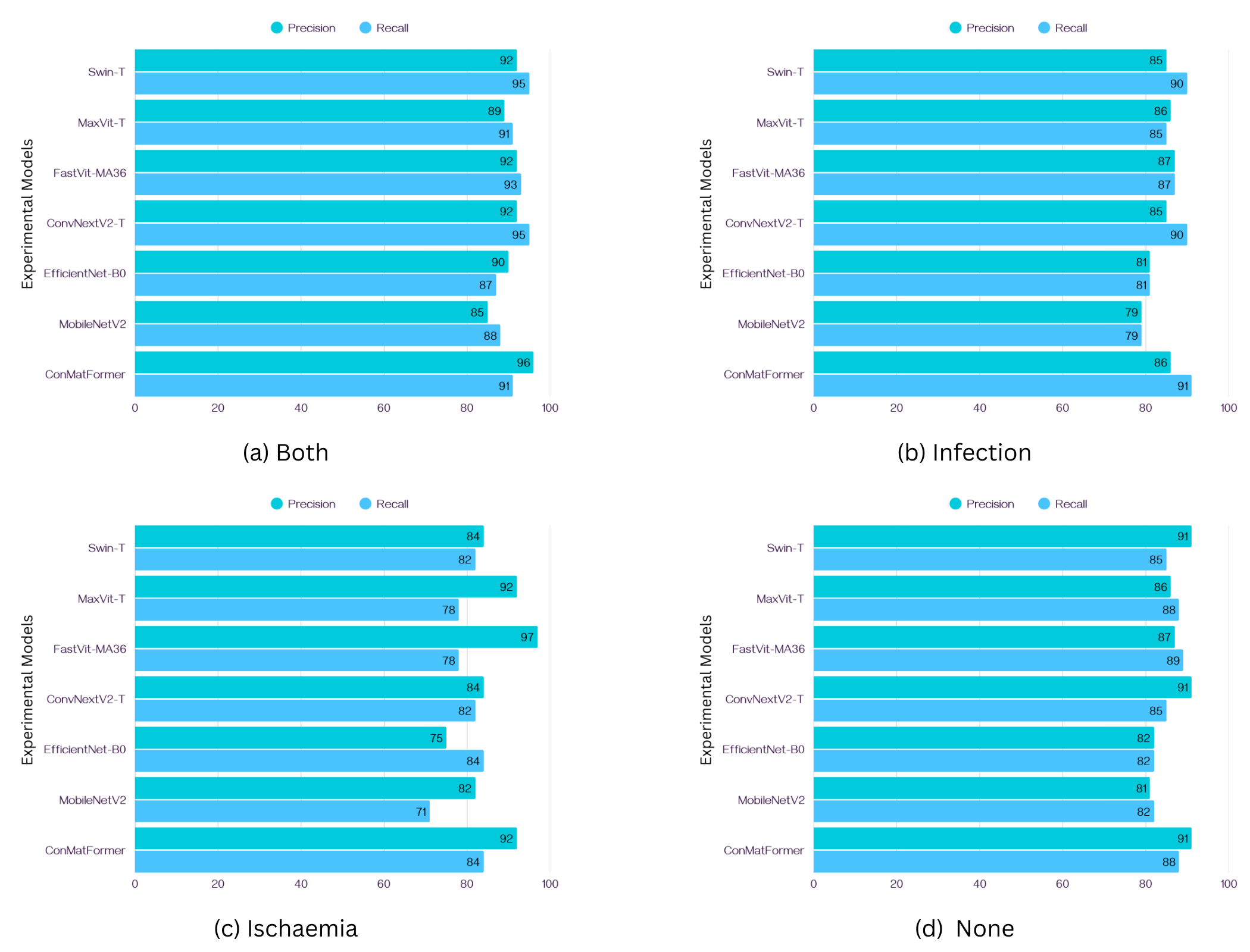}
    \caption{class-wise Recall and precision comparison of various models for DFU classification on DS1.}
    \label{fig:class-wise Recall and precision}
\end{figure}
In the ``both” class, ConMatFormer led, with both precision and recall reaching 0.96, suggesting its strong consistency in handling complex dual-condition cases. Swin-T and ConvNextV2-T also performed well in this class, each maintaining high values close to ConMatFormer. ConMatFormer once again shows competitive scores in the ``infection” category, though slightly edged out in recall by FastViT-MA36.

FastViTMA36 had the strongest recall (0.97) when concentrating on "ischemia," but its precision (0.78) decreased. With both values above 0.88, ConMatFormer continued to perform more evenly. With scores often falling between 0.82 and 0.91, the "none" class displayed comparatively consistent results across all models. In almost every category, lightweight models like MobileNetV2 and EfficientNet-B0 performed worse, particularly in the precision of ischemia. They trail behind the best-performing models by more than 0.100.15. As seen in Figure ~\ref{fig:class-wise Recall and precision}, ConMatFormer was the most well-rounded model overall, with good precision and recall in every diagnostic category. 

\begin{figure}[h]
    \centering
    \includegraphics[width=0.8\linewidth]{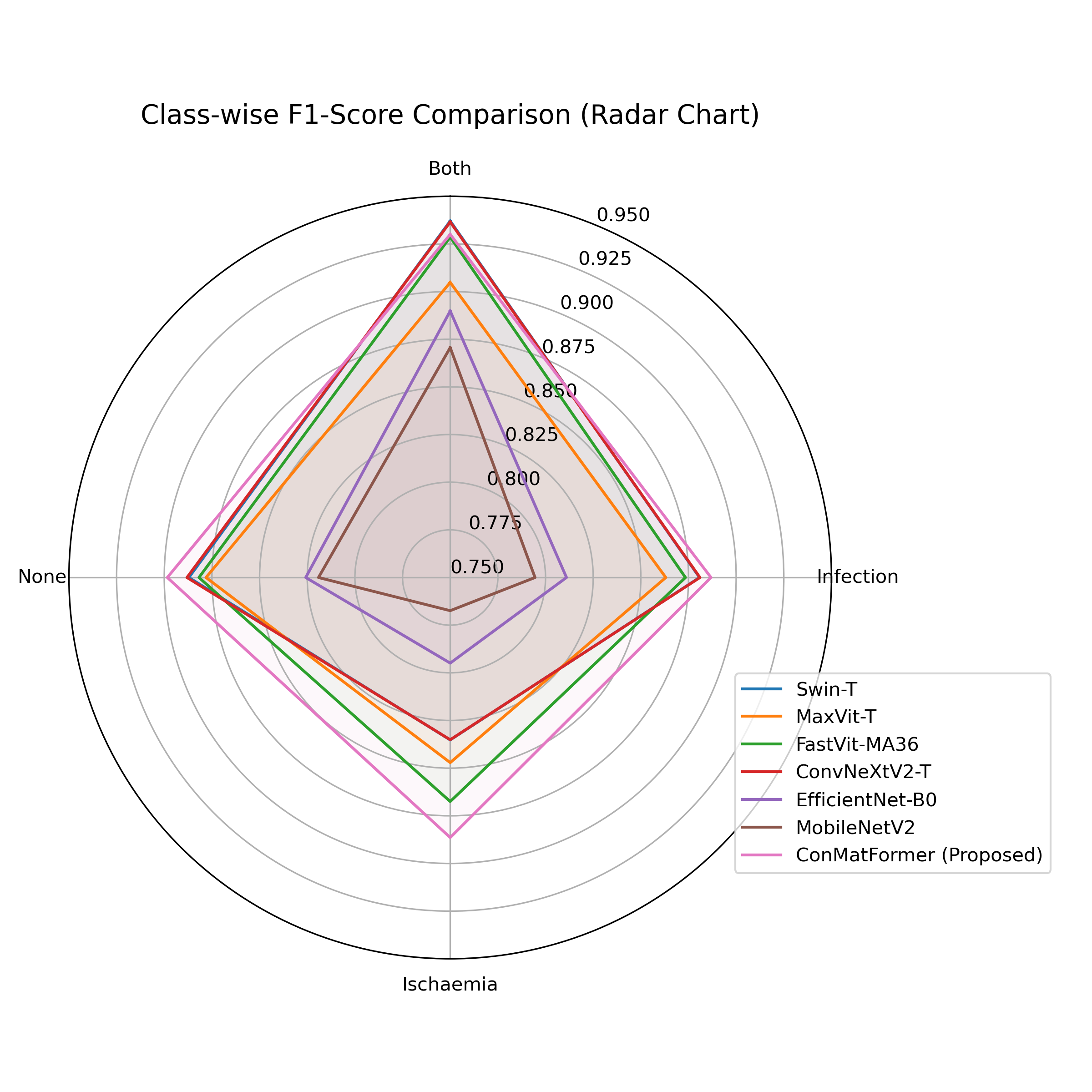}
    \caption{Radar chart illustrating class-wise F1-score comparison of various models for DFU classification on DS1.}
    \label{fig:classwise-f1-radar}
\end{figure}

The suggested ConMatFormer shown constant high performance across all four classes, according to the class-wise F1-score. Both were 0.9300, infection was 0.8868, ischemia was 0.8864, and none was 0.8984. With F1-scores of 0.9370 and 0.9365, respectively, which were marginally higher than those of the suggested model, Swin-T and ConvNextV2-T outperformed the other models in “both” class.

However, their performance decreased in other categories, such as ischaemia, where both remained at approximately 0.835. FastVit-MA36 offered more balanced results, scoring above 0.86 in all classes. MaxVit-T maintained a stable performance but did not surpass the others in any class. Efficient-B0 and MobileNetV2 performed noticeably worse, with their ischaemia and infection scores falling below 0.82 and, in some cases, near 0.76, respectively. These comparisons highlight that ConMatFormer delivers a more even and higher performance across all classes, especially in challenging categories such as ischaemia, as shown in Figure ~\ref{fig:classwise-f1-radar}.\\
\begin{figure}[htbp]
    \centering
    \includegraphics[width=0.9\textwidth]{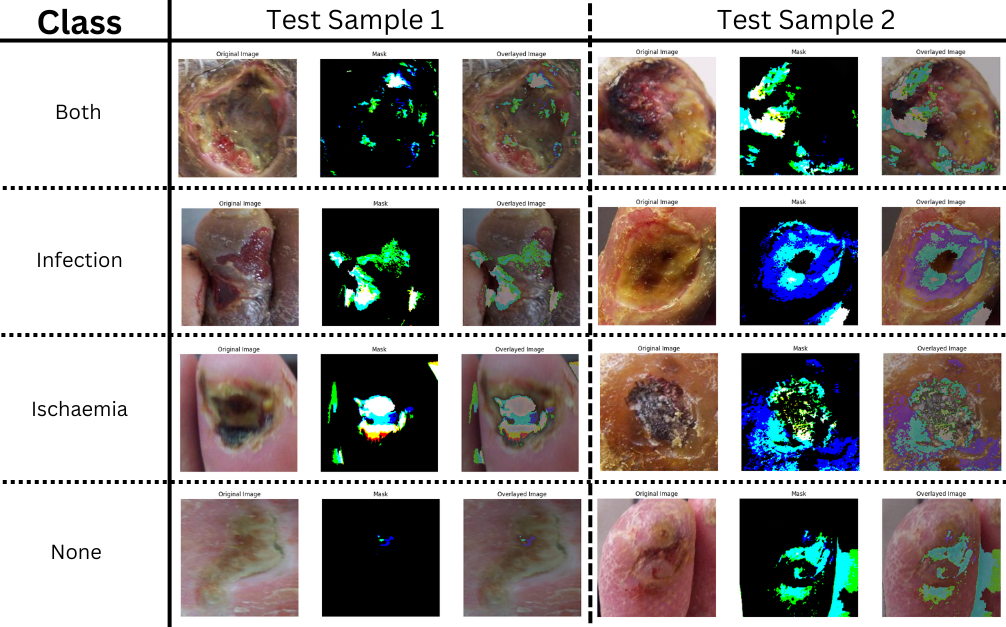}
    \caption{Lime Explanation for Proposed ConMatFormer Model used in this Study on DS1.}
    \label{fig:dfu_lime}
\end{figure}

Figure~\ref{fig:dfu_lime} exhibits the LIME-based visual explanations for the proposed ConMatFormer model's predictions for four types of DFUs: both (infection and ischaemia), infection, ischaemia, and none. Two test samples were used for each class. Each sample has the original image, LIME-generated saliency mask, and an overlaid image that combines the two. The saliency masks show which parts of the image had the biggest effect on the model's decision. LIME accurately identifies clinically important features, such as discolored tissue, necrotic areas, and ulcer borders in cases of infection and ischaemia. This shows that the model focuses on meaningful dermatological patterns. For the ``none" class, a small amount of activation in the mask showed that there were no features related to ulcers, which made the model more reliable in negative cases. These visualizations show that ConMatFormer is easy to understand and confirm that it can be safely used in clinical settings by aligning model attention with areas that are important for pathology.

\begin{figure}[htbp]
    \centering
    \includegraphics[width=0.9\textwidth]{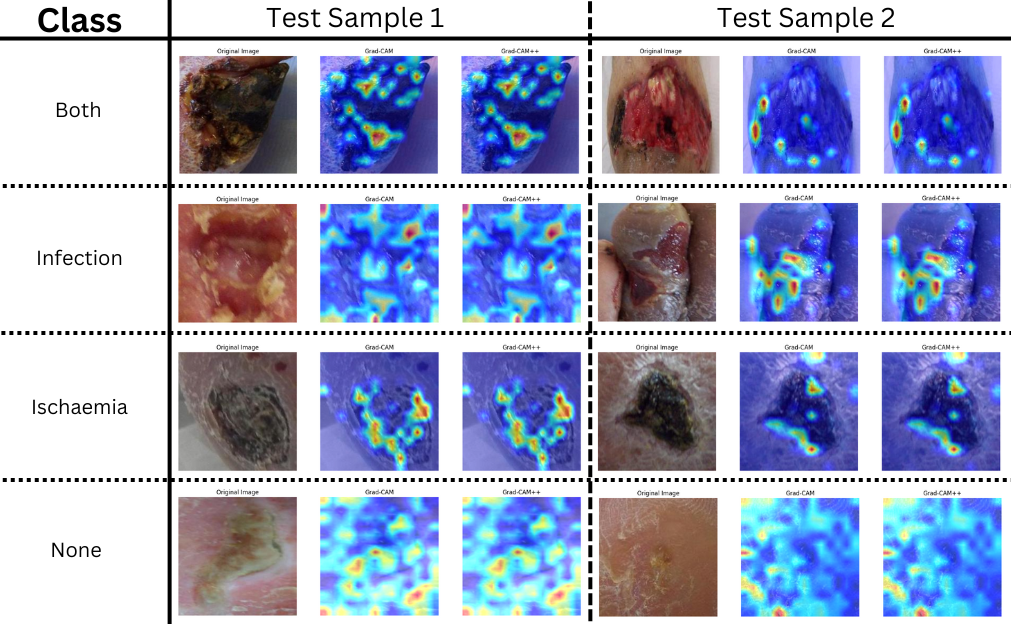}
    \caption{Grad-Cam and Grad-Cam++ Visualization for Proposed ConMatFormer Model used in this Study on DS1.}
    \label{fig:dfu_gdcam}
\end{figure}

Figure~\ref{fig:dfu_gdcam} presents the Grad-CAM and Grad-CAM++ visual explanations for the ConMatFormer model's predictions for the four types of DFUs: both, infection, ischaemia, and none. There is one class per row, and each column shows two test samples (test samples 1 and 2). Each sample has an original image, Grad-CAM heatmap, and Grad-CAM++ heatmap. The GradCAM and GradCAM++ heatmaps show areas of space that have a significant effect on the model prediction. Areas of high importance are shown in warmer colors, such as red and yellow. In the ``both",``infection" and ``ischaemia" classes, these visualizations always point to medically important ulcer areas, such as necrotic tissue, inflammation zones, and ischemic skin. This shows that the model's decisions were consistent with clinical pathology. 

Grad-CAM++ is more adept than standard Grad-CAM at picking out minute details, particularly when picking out delicate characteristics in intricate ulcer patterns. Both Grad-CAM and Grad-CAM++ display diffuse and low-activation heatmaps for the "none" class. This demonstrates that the model accurately and without deceptive focus detected skin that was not ulcerated. All things considered, this figure demonstrates that the ConMatFormer model is dependable and simple to comprehend, and that its predictions are grounded in imaging characteristics that are clinically significant.

Across the four DFU classes, the Grad-CAM model fared better than LIME in terms of offering concise and clinically significant visual explanations for the ConMatFormer model. Grad-CAM is easy to use and interpret for clinical validation since it continuously uses smooth and continuous heatmaps to highlight pertinent lesion locations. On the contrary, LIME has the ability of generating sparse or noisy masks and can further exhibit greater variability among samples, especially in case of infection and ischemia. Due to the inconsistency in their performance, the overlayed masks do not seem to be a reliable choice in clinical settings. As such, GradCAM is a good choice for elucidating DFU classification outcomes in medical image analysis since it provides more robust attention maps.

\begin{figure}[htbp]
    \centering
    \includegraphics[width=0.9\textwidth]{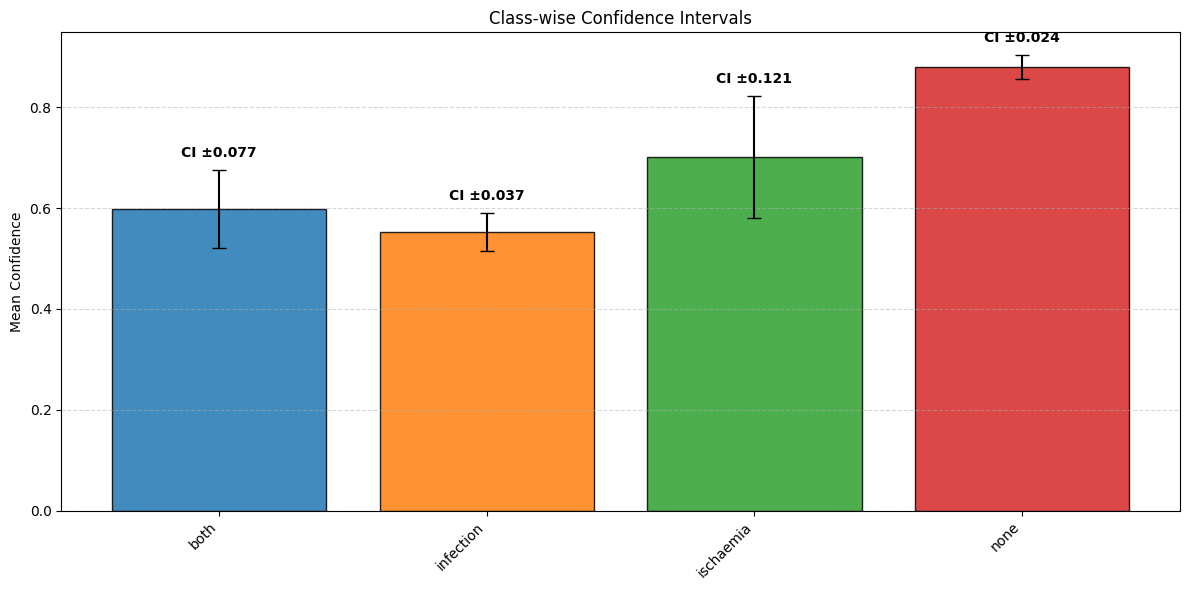}
    \caption{Classwise Confidence Interval of the Proposed ConMatFormer Model used in this Study on DS1.}
    \label{fig:dfu_ci}
\end{figure}
Figure \ref{fig:dfu_ci} illustrates the confidence intervals for each category of predictions generated by the ConMatFormer model on the test dataset. The model displayed the most confidence and the least amount of ambiguity for the ‘none’ category with a mean of 0.89 and a confidence range of $\pm 0.024$. The categories of "ischemia," "both," and "infection" were evaluated next. Interestingly, the "ischemia" category showed the broadest confidence range $(\pm 0.121)$, suggesting that the model's certainty varied most. These results imply the fact that ConMatFormer is good in detecting zero cases, but predictions for particular illnesses show greater variability, either as a result of overlapping visual characteristics or a lack of sample diversity in these groups.

\subsection{Evaluation with Cross Validation}
To provide statistical validation of our results, we further evaluated the proposed model using four-fold cross-validation (4-fold CV), reporting the mean ± standard deviation across folds and conducting paired t-tests against the baselines.

\begin{table}[h]
\centering
\caption{Cross-validation performance comparison of different models. Values are reported as mean $\pm$ standard deviation across 4 folds on DS1.}
\label{tab:comparison}
\begin{tabular}{lcccc}
\hline
\textbf{Model} & \textbf{Accuracy} & \textbf{Precision} & \textbf{Recall} & \textbf{F1-score} \\
\hline
Swin-T  & 0.9605 $\pm$ 0.0057 & 0.9614 $\pm$ 0.0053 & 0.9613 $\pm$ 0.0055 & 0.9612 $\pm$ 0.0055 \\
MaxVit-T          & 0.9375 $\pm$ 0.0060 & 0.9384 $\pm$ 0.0059 & 0.9389 $\pm$ 0.0059 & 0.9385 $\pm$ 0.0059 \\
FastVit-MA36          & 0.9712 $\pm$ 0.0068 & 0.9721 $\pm$ 0.0060 & 0.9718 $\pm$ 0.0066 & 0.9717 $\pm$ 0.0066 \\
ConvNeXtV2-T         & 0.9577 $\pm$ 0.0042 & 0.9580 $\pm$ 0.0043 & 0.9575 $\pm$ 0.0042 & 0.9577 $\pm$ 0.0042 \\
EfficientNet-B0   & 0.8940 $\pm$ 0.0078 & 0.8967 $\pm$ 0.0061 & 0.8962 $\pm$ 0.0077 & 0.8958 $\pm$ 0.0075 \\
MobileNetV2      & 0.9143 $\pm$ 0.0081 & 0.9165 $\pm$ 0.0083 & 0.9161 $\pm$ 0.0079 & 0.9160 $\pm$ 0.0078 \\
\textbf{ConMatFormer (Proposed)} & \textbf{0.9755 $\pm$ 0.0031} & \textbf{0.9759 $\pm$ 0.0031} & \textbf{0.9760 $\pm$ 0.0030} & \textbf{0.9759 $\pm$ 0.0031} \\
\hline
\end{tabular}
\end{table}

\begin{table}[h]
\centering
\caption{Paired t-test results (fold-wise accuracy) comparing the proposed ConMatFormer against baseline models. Statistically significant differences are marked with $^\ast$ (p < 0.05).}
\label{tab:ttest_accuracy}
\begin{tabular}{lccc}
\hline
\textbf{Comparison (Proposed vs.)} & \textbf{t-statistic} & \textbf{p-value} & \textbf{Significant?} \\
\hline
Swin-T  & 6.504  & 0.0074   & Yes$^\ast$ \\
MaxVit-T         & 18.058 & 0.00037  & Yes$^\ast$ \\
FastVit-MA36           & 1.218  & 0.3103   & No \\
ConvNeXtV2-T         & 4.852  & 0.0167   & Yes$^\ast$ \\
EfficientNet-B0    & 26.449 & 0.00012  & Yes$^\ast$ \\
MobileNetV2      & 23.388 & 0.00017  & Yes$^\ast$ \\
\hline
\end{tabular}
\end{table}
\noindent As summarized in Table~\ref{tab:comparison}, the proposed \textbf{ConMatFormer} achieves the highest mean performance on DS1 under 4-fold cross-validation— \(\text{Acc}=0.9755\pm0.0031\), \(\text{Prec}=0.9759\pm0.0031\), \(\text{Rec}=0.9760\pm0.0030\), and \(\text{F1}=0.9759\pm0.0031\)—with lower variance than most baselines. In accuracy, this yields absolute gains of \(+1.50\) percentage points (pp) over Swin-T (0.9605), \(+1.78\) pp over ConvNeXtV2-T (0.9577), \(+6.12\) pp over MobileNetV2 (0.9143), and \(+8.15\) pp over EfficientNet-B0 (0.8940); FastViT-MA36 is the closest competitor with a smaller gap of \(+0.43\) pp (0.9712).

\noindent To establish statistical significance, we conducted two-sided paired \(t\)-tests on fold-wise accuracies at \(\alpha=0.05\) (Table~\ref{tab:ttest_accuracy}). Improvements over Swin-T, MaxViT-T, ConvNeXtV2-T, EfficientNet-B0, and MobileNetV2 are \emph{statistically significant} (\(p<0.05\); marked with \(^{\ast}\)), whereas the difference with FastViT-MA36 is not significant (\(p=0.3103\)). Collectively, Tables~\ref{tab:comparison} and~\ref{tab:ttest_accuracy} indicate that combining ConvNeXt features with CBAM, DANet, and a transformer branch yields both higher accuracy and greater stability than CNN-only or transformer-only backbones on DS1.

\begin{figure}[htbp]
    \centering
    \includegraphics[width=0.9\textwidth]{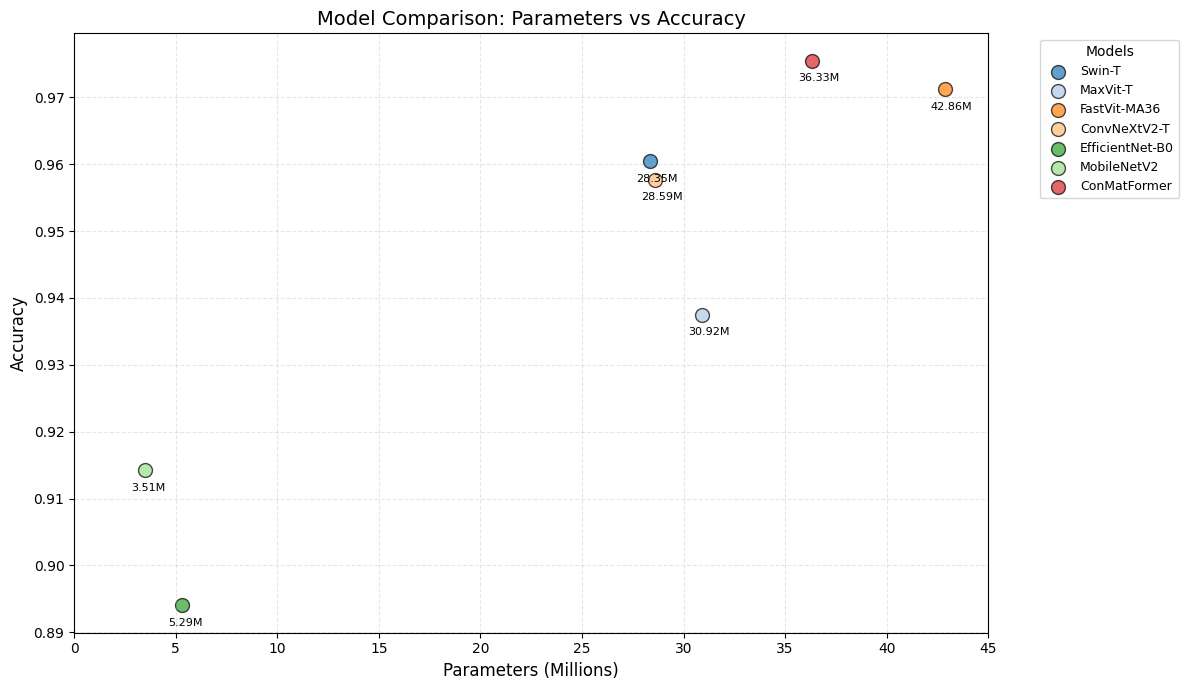}
    \caption{ConMatFormers' excellent combination of complexity and performance is highlighted, which compares model correctness against parameter size.}
    \label{fig:accprm}
\end{figure}

 A comparison of several architectures in terms of parameter sizes is shown in Figure \ref{fig:accprm}. Although they needed a lot of processing power, models with more parameters, like FastViT-MA36 (42.86M, 0.9712), showed remarkable accuracy. On the other hand, smaller models with lower accuracy, like EfficientNet-B0 (5.29M, 0.8940) and MobileNetV2 (3.51M, 0.9143), showed higher efficiency. Notably, our suggested ConMatFormer (36.33M, 0.9755) outperformed even the models with more parameters, achieving the highest accuracy of all the models tested. This shows that it can provide state-of-the-art performance with a manageable parameter size when compared to other models. This equilibrium confirms that ConMatFormer is accurate and flexible for real-world uses.

\begin{figure}[htbp]
    \centering
    \includegraphics[width=0.9\textwidth]{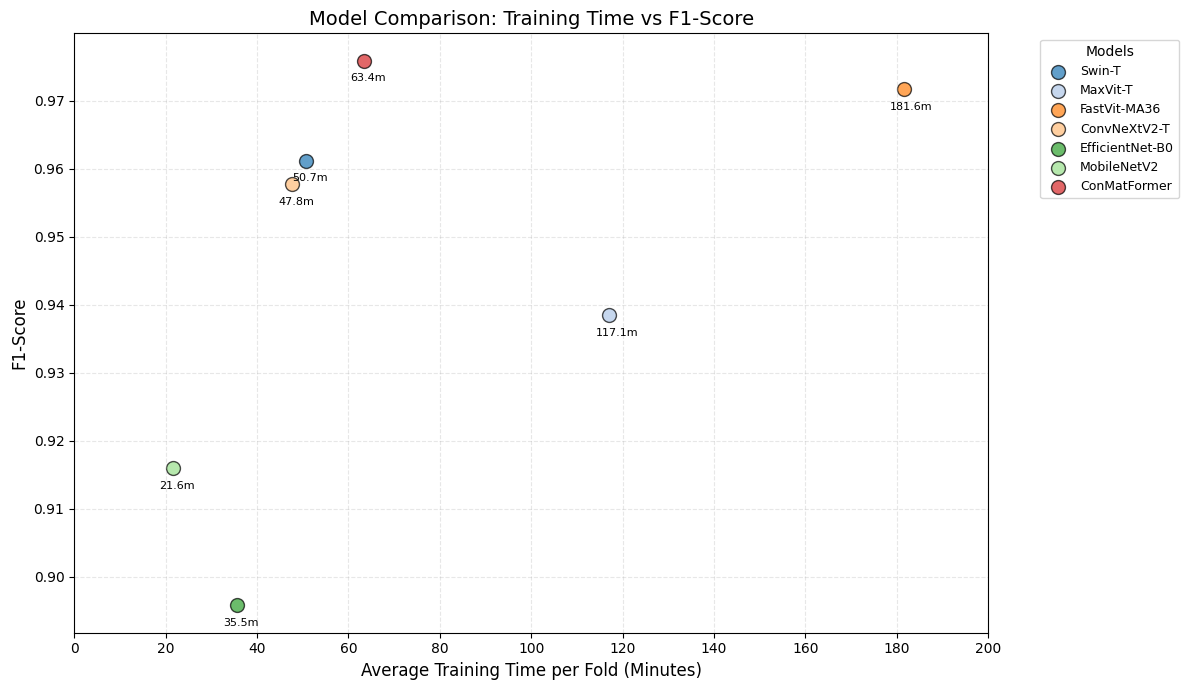}
    \caption{The balance between prediction reliability and efficiency provided by Comparing training time to F1-score of ConMatFormer Model.}
    \label{fig:timef1}
\end{figure}

 The correlation between predicting performance and training efficiency is depicted in Figure \ref{fig:timef1}. Notably, the suggested ConMatFormer model outperformed FastViT-MA36, which needed the greatest training time (181.56 min) to reach an F1-score of 0.9717. With just 63.44 minutes of training per fold, ConMatFormer achieved the highest F1-score of 0.9759, proving that it can provide greater predictive reliability with significantly less computing effort than larger models. The F1-scores of smaller models, including MobileNetV2 (21.65 min, 0.9160) and EfficientNet-B0 (35.55 min, 0.8958), were noticeably lower despite being trained more quickly. A balanced result is provided by mid-range models such as ConvNeXtV2-T (47.75 min, 0.9577) and Swin-T (50.67 min, 0.9612). Nonetheless, ConMatFormer's steady improvement demonstrates its effectiveness as a well-tuned model that offers strong generalization and excellent accuracy at a moderate training time.

\subsection{Ablation Study}
To isolate the contribution of each architectural component, we conducted a systematic ablation on the DFU 2021 dataset. Table \ref{tab:Performance comparison of baseline and attention-based models on the DFU 2021 dataset} illustrates the performance of the deep learning models and attention mechanisms used for the classification of the DFU 2021 dataset. The performance of ConvNeXtV2-T, CBAM, DANet, and CBAM+DANet is compared with that of the proposed model, which is an amalgamation of CBAM, DANet, and Transformer. We calculated the accuracy, precision, recall, and F1-score of all these models against those of our proposed model. As shown in Table \ref{tab:Performance comparison of baseline and attention-based models on the DFU 2021 dataset} the baseline ConvNeXtV2-T performs well across all metrics, whereas CBAM captures more true positives but reduces the precision and accuracy. DANet provided a balanced but weaker performance than CBAM, whereas the hybrid of these two models output better precision. Our proposed model achieved the best overall balance with the highest accuracy of 0.8961 and best F1-score of 0.9004, indicating that adding a transformer helps the model leverage both local and global (attention) features.
\begin{table}[h]
\centering
\caption{Performance comparison of baseline and attention-based models on the DS1 dataset.}
\label{tab:Performance comparison of baseline and attention-based models on the DFU 2021 dataset}
\begin{tabular}{|l|c|c|c|c|}
\hline
\textbf{Component / Model} & \textbf{Accuracy} & \textbf{Precision } & \textbf{Recall } & \textbf{F1-score } \\
\hline
ConvNeXtV2-T                & 0.8860 & 0.8841 & 0.8848 & 0.8840 \\
ConvNeXtV2-T+CBAM                        & 0.8650 & 0.8755 & 0.8909 & 0.8829 \\
ConvNeXtV2-T+DANet                       & 0.8734 & 0.8758 & 0.8673 & 0.8709 \\
ConvNeXtV2-T+CBAM+DANet                  & 0.8768 & 0.9017 & 0.8472 & 0.8693 \\
ConvNeXtV2-T+CBAM+DANet+Transformer (Proposed) & 0.8961 & 0.9160 & 0.8866 & 0.9004 \\
\hline
\end{tabular}
\end{table}

\begin{table}[H]
\centering
\caption{A methodical ablation on the DFU 2021 dataset was conducted in order to separate the influence of each architectural element.}
\label{tab:Comparison of classification performance across different experiments on DFU 2021 dataset}
\begin{tabular}{|l|l|c|c|c|c|}
\hline
\textbf{Experiment} & \textbf{Class} & \textbf{Precision} & \textbf{Recall} & \textbf{F1-score} & \textbf{Accuracy} \\
\hline
\multirow{2}{*}{Ischaemia vs None} 
  & Ischaemia & 0.9756 & 0.8696 & 0.9195 & \multirow{2}{*}{0.9874} \\
  & None      & 0.9884 & 0.9980 & 0.9932 &  \\
\hline
\multirow{2}{*}{Infection vs None} 
  & Infection & 0.8864 & 0.9159 & 0.9009 & \multirow{2}{*}{0.8992} \\
  & None      & 0.9130 & 0.8826 & 0.8975 &  \\
\hline
\multirow{2}{*}{Both vs None} 
  & Both      & 0.9835 & 0.9520 & 0.9675 & \multirow{2}{*}{0.9874} \\
  & None      & 0.9883 & 0.9961 & 0.9922 &  \\
\hline
\multirow{3}{*}{3-Class (Infection + Ischaemia + None)} 
  & Infection & 0.8774 & 0.8963 & 0.8867 & \multirow{3}{*}{0.8876} \\
  & Ischaemia & 0.9268 & 0.8261 & 0.8736 &  \\
  & None      & 0.8950 & 0.8845 & 0.8898 &  \\
\hline
\end{tabular}
\end{table}

Table \ref{tab:Comparison of classification performance across different experiments on DFU 2021 dataset} presents a comparative analysis of the classification performance under various experimental conditions using the DS1 dataset. With both the Ischemia vs. None and Both vs. None tasks attaining an overall accuracy of 0.9874, backed by good precision (varying from 0.9835 to 0.9888) and F1-scores exceeding 0.96, the binary classification tasks showed impressive results. Somewhat lower but balanced recall and precision was seen in the None and Infection class which further demonstrated an accuracy of 0.8992 in the Infection vs. None challenge. In contrast to the binary tasks, the more intricate 3-Class classification (infection, ischemia, and none) produced a lower accuracy of 0.8876 along with lower precision and recall for individual classes. According to these results, the inclusion of several categories inside a single framework introduces complexity and results in a small loss in accuracy and F1-scores, even though the binary distinction of sick circumstances from healthy cases produces greater performance. In addition to highlighting the difficulties involved in simultaneously classifying different disorders in medical image analysis, this ablation study emphasises the trade-off between binary and multi-class techniques.

\section{Discussion}

\begin{table}[h]
\centering
\caption{Performance Comparison of ConMatFormer with State-of-the-Art Models on DFUC2021 and DFU Datasets}
\begin{tabular}{|>{\raggedright\arraybackslash}p{2cm} 
                |>{\raggedright\arraybackslash}p{3cm} 
                |>{\raggedright\arraybackslash}p{2.8cm} 
                |>{\raggedright\arraybackslash}p{3.8cm} 
                |>{\centering\arraybackslash}p{2cm}|}
\hline
\textbf{Literature} & \textbf{Method} & \textbf{Dataset} & \textbf{Classes} & \textbf{Accuracy} \\
\hline
\cite{10} & Dual-track architecture: Swin Transformer + EMADN & DFUC-2021 & Infection, Ischaemia, None, Both & 78.79\% \\
\hline
\cite{11} & Dense-ShuffleGCN ANet & DFUC-2021 & Infection, Ischaemia, None, Both & 86.00\% \\
\hline
\cite{16} & CNNs (EfficientNet, ResNet), ViTs (ViT, DeiT), optimized with SAM & DFUC-2021 & Control, Infection, Ischaemia, Both & 88.49\% \\
\hline
\cite{12} & AlexNet, VGG16/19, GoogLeNet, ResNet50/101, MobileNet, SqueezeNet, DenseNet & DFUC2020 & Ischaemia: Positive/Negative; Infection: Positive/Negative & 99.49\% (Ischaemia), 84.76\% (Infection) \\
\hline
\cite{14} & DFU\_SPNet & DFU & Normal Skin, Abnormal Skin & 97.40\% \\
\hline
\cite{23} & EfficientNet, AlexNet, GoogLeNet, VGG16, VGG19 & DFU & Normal, Abnormal & 98.97\% \\
\hline
\cite{fadhel2024} & DFU\_TFNet (Transfer Learning) & DFU & Normal, Abnormal & 99.8\% \\
\hline
\textbf{This Study} & \textbf{ConMatFormer} & \textbf{DFUC-2021, DFU} & \textbf{Infection, Ischaemia, None, Both; Ulcer vs Healthy} & \textbf{89.61\%, 99.50\%}\\
\hline
\end{tabular}
\label{tab:model_performance_comparison}
\end{table}

The purpose of this study was to determine whether it is possible to create a DL framework that successfully combines the advantages of transformer-based and convolutional architectures while achieving SOTA accuracy. In order to solve this issue, our ConvMatformer approach uses transformers with self-attention for the global contextual information of images and convolution for local features. We used two publicly available benchmark datasets, DS1 (DFUC2021) and DS2 (diabetic foot ulcer (DFU)), to assess ConvFormer's functionality and performance. A fair comparison with the SOTA techniques is ensured by this procedure. ConvMatformer was also assessed against six other DL techniques that are currently in use: Swin-T, MaxVit-T, FastVit-MA36, ConvNeXtV2-T, EfficientNet-B0, and MobileNetV2. The ConvMatformer continuously beat these models in terms of accuracy, recall, precision, F1-score, and AUC, as indicated in Table~\ref{tab:model_performance_comparison}. Our architecture included XAI approaches to guarantee openness and confidence in healthcare applications. ConvMatformer's decision-making process was visualised and interpreted using Grad-CAM and LIME. These illustrations show that the algorithm continuously concentrates on areas of the input images that are important to medicine. In conclusion, our findings demonstrate that a composite structure, as ConvMatformer, can effectively recognise both minor and major details, function well, and be clearly comprehended. Although the performance evaluation is the major focus of this study, further statistical studies contrasting ConvMatformer with baseline models are still a topic for further investigation.

\subsection{Clinical Adoption Pathway}

Both technical performance and clinical usability must be guaranteed for artificial intelligence models, like ConMatFormer, to be successfully incorporated into actual medical practice. There are multiple ways in which the suggested framework could serve as a decision-support tool. In order to confirm whether the model is concentrating on clinically relevant aspects, doctors can first visualise model attention regions (such as ischemia zones, infection boundaries, or ulcer regions) by using explainability tools like Grad-CAM, Grad-CAM++, and LIME. Second, especially in cases that are unclear, the system might offer a second opinion to help doctors confirm diagnosis. Thirdly, by giving priority to patients that need immediate attention, it could act as a triage assistant in healthcare settings with limited resources. Lastly, by demonstrating the decision-making cues, it can serve as an instructional tool for junior clinicians and medical students. From an integration standpoint, ConMatFormer's user-friendly interface, which combines predictions with visual explanations, allows it to be integrated into hospital infrastructures already in place, such as electronic health records (EHR) and picture archiving and communication systems (PACS). 
Crucially, in order to improve user trust, correct possible misclassifications, and improve system outputs, clinician input must be consistently integrated into iterative design cycles. This method emphasises that clinical adoption depends on interpretability, usability, and cooperation with medical experts in addition to accurate measures.

\section{Limitations and Future Work}
Despite its great accuracy and F1-score, the ConMatFormer model has some limitations:
\begin{itemize}
    \item First, to address the risk of overfitting, we employed weight decay, dropout layers, data augmentation, and cross-validation. We cannot, however, totally rule out the chance of overfitting. The model must be trained on bigger and more varied datasets in order to test its generalisation further.

    \item Second, only DFU datasets that were gathered, made publicly available, and anonymised (DFUC2021 and DFU) were used to train and test the model. Despite being frequently used as standards, these datasets could not accurately capture the variety that exists in real-world clinical practice (e.g., variances in acquisition equipment, clinical surroundings, and ethnicity). Because of this reliance on the datasets gathered, we are unable to claim universal applicability. Future studies ought to look into domain adaptation techniques and multicenter datasets.

    \item Third, we did not thoroughly examine factors like energy use, inference time, and memory usage. Even though our design strikes a balance between efficiency and accuracy, a more detailed analysis of the computational footprint is required to assess its viability in real-time clinical settings. 
    \item Finally, fall of the datasets used were publicly available and anonymized, so no additional ethical approval was required. 
\end{itemize}

In the future, there are many promising research paths that could make the ConMatFormer even more robust and useful in the clinic.
\begin{itemize}
    \item Investigating ways to optimise the architecture in order to lower energy use, speed up inference, and reduce memory utilization is very crucial. These modifications aid to enhance the framework for use in portable clinical field devices as well as in real-time hospital systems, where hardware resources are frequently scarce. 
    \item Incorporating multi-modal integration is beneficial and necessary. Adding thermal imaging data to standard visual scans tends to enhance classification performance in cases where abnormal temperature patterns frequently occur before the development of visible ulcers. Integrating optical and thermographic characteristics seem to aid in the earlier and more accurate detection of DFUs.
    \item Testing the model on a range of clinical contexts and medical datasets is salient to ensure the model’s general functionality. ConMatFormer's performance with various patient types, imaging situations, and healthcare settings might be ascertained by testing it on other DFU datasets as well as other ulcer or wound imaging datasets.
\end{itemize}

These guidelines will assist ConMatFormer in developing into a clinically valuable, scalable, and effective diabetes diagnostic tool.
\section{Conclusion}
Millions of people worldwide suffer from diabetic foot ulcers (DFU), a serious health issue that can lead to serious consequences like amputations. Due to the complex and varied nature of DFU, early and precise detection is essential for effective management and prompt intervention. Nevertheless, this is a difficult problem. The healthcare system and medical personnel experience major delays as a result of the time-consuming, subjective, and resource-intensive nature of conventional diagnostic techniques. By employing a transparent deep learning model for the automatic classification of DFU, our study gets around these challenges. The suggested model efficiently captures both broad and detailed aspects at the same time by utilising 2D ConvNext Blocks, a parallel transformer module, and a multi-attention mechanism. Following feature concatenation, the discriminative regions were attended to using the CBAM and DANet attention techniques. ConvNext blocks use global response normalisation to collect data from several model levels. The combination of attention and ConvNext-based transformers improves the ability to extract features and generalise models. By adding the DANet module, the network may selectively concentrate on pertinent details without adding to its computing load, increasing accuracy.
Finally, the model was tested on the DFUC-2021 and DFU datasets, achieving a classification accuracy of 0.8961 and a macro F1-score of 0.9004 on the DS1 dataset, and an accuracy of 0.9950 and a macro F1-score of 0.9950 on the DS2 dataset. A comparison with other state-of-the-art DL models and previous studies showed that the proposed network exhibited better performance for DFU classification. Moreover, LIME and Grad-CAM were used to generate visual interpretations, which helped identify the regions of interest in an image from the model’s perspective. This XAI analysis offers a transparent and explainable approach for the practical use of the model in clinical settings.

\section*{Data Availability}
Used datasets "Diabetic Foot Ulcer (DFU-2021)" and "Kaggle DFU" can be accessed by the link:
\url{https://dfu-2021.grand-challenge.org/Dataset/} and
\\
\url{https://www.kaggle.com/datasets/laithjj/diabetic-foot-ulcer-dfu}


 \section*{Funding}
No funding was received for conducting this research or for preparing this manuscript.

\section*{Declaration of Competing Interest}
The authors declare that they have no known competing financial interests or personal relationships that could influence the work reported in this study.



\end{document}